\documentclass[10pt,twocolumn,letterpaper]{article}

\usepackage[pagenumbers]{cvpr} % To force page numbers, e.g. for an arXiv version

\usepackage{graphicx}
\usepackage{amsmath}
\usepackage{amssymb}
\usepackage{booktabs}
\usepackage{makecell}

\usepackage[pagebackref,breaklinks,colorlinks]{hyperref}
\hypersetup{
    colorlinks=true,
    linkcolor=blue,
    urlcolor=blue,
}
\usepackage{stackrel,amsmath,amsfonts,enumitem,booktabs,multirow,caption,cite}

\usepackage{xcolor,colortbl}
\definecolor{Gray}{gray}{0.9}

\definecolor{roadcolor}{rgb}{0.5, 0.25, 0.5}
\definecolor{sidewalkcolor}{rgb}{0.96, 0.14, 0.91}
\definecolor{buildingcolor}{rgb}{0.27, 0.27, 0.27}
\definecolor{wallcolor}{rgb}{0.4, 0.4, 0.61}
\definecolor{fencecolor}{rgb}{0.75, 0.6, 0.6}
\definecolor{polecolor}{rgb}{0.6, 0.6, 0.6}
\definecolor{trafficlightcolor}{rgb}{0.98, 0.67, 0.12}
\definecolor{trafficsigncolor}{rgb}{0.86, 0.86, 0}
\definecolor{vegetationcolor}{rgb}{0.42, 0.56, 0.14}
\definecolor{terraincolor}{rgb}{0.6, 0.98, 0.6}
\definecolor{skycolor}{rgb}{0.27, 0.51, 0.71}
\definecolor{personcolor}{rgb}{0.86, 0.08, 0.24}
\definecolor{ridercolor}{rgb}{1, 0, 0}
\definecolor{carcolor}{rgb}{0, 0, 0.56}
\definecolor{truckcolor}{rgb}{0, 0, 0.27}
\definecolor{buscolor}{rgb}{0, 0.24, 0.39}
\definecolor{motorcyclecolor}{rgb}{0, 0, 0.90}
\definecolor{bicyclecolor}{rgb}{0.47, 0.04, 0.13}

\usepackage[capitalize]{cleveref}
\crefname{section}{Sec.}{Secs.}
\Crefname{section}{Section}{Sections}
\Crefname{table}{Table}{Tables}
\crefname{table}{Tab.}{Tabs.}

\def\cvprPaperID{3020} % *** Enter the CVPR Paper ID here
\def\confName{CVPR}
\def\confYear{2022}
\def\ie{\textit{i.e.}}

\def\model{F}
\def\sty{\text{sty}}
\def\fog{\text{fog}}
\def\dual{\text{dual}}
\def\imd{m}
\def\td{t}
\newcommand{\heading}[1]{\noindent\textbf{#1}}

\usepackage{pifont}
\newcommand{\cmark}{\ding{51}}%

\begin{document}

\title{Both Style and Fog Matter:\\Cumulative Domain Adaptation for Semantic Foggy Scene Understanding}

\author{Xianzheng Ma$^1$, \hspace{1mm}Zhixiang Wang$^2$, \hspace{1mm}Yacheng Zhan$^1$, \hspace{1mm}Yinqiang Zheng$^2$ , \hspace{1mm}Zheng Wang$^1$, \\Dengxin Dai$^3$, \hspace{1mm}Chia-Wen Lin$^4$\\
$^1$Wuhan University  \hspace{3mm}  $^2$UTokyo  \hspace{3mm}  $^3$MPI for Informatics  \hspace{3mm}  $^4$NTHU\\

% {\tt\small maxianzheng@whu.edu.cn}

% \and
}
\maketitle

%%%%%%%%% ABSTRACT
\begin{abstract}
Although considerable progress has been made in semantic scene understanding under clear weather, it is still a tough problem under adverse weather conditions, such as dense fog, due to the uncertainty caused by imperfect observations. Besides, difficulties in collecting and labeling foggy images hinder the progress of this field.
Considering the success in semantic scene understanding under clear weather, we think it is reasonable to transfer knowledge learned from clear images to the foggy domain. As such, the problem becomes to bridge the domain gap between clear images and foggy images. 
Unlike previous methods that mainly focus on closing the domain gap caused by fog --- defogging the foggy images or fogging the clear images, we propose to alleviate the domain gap by considering fog influence and style variation simultaneously. 
The motivation is based on our finding that the style-related gap and the fog-related gap can be divided and closed respectively, by adding an intermediate domain.
Thus, we propose a new pipeline to cumulatively adapt style, fog and the dual-factor (style and fog). Specifically, we devise a unified framework to disentangle the style factor and the fog factor separately, and then the dual-factor from images in different domains. 
Furthermore, we collaborate the disentanglement of three factors with a novel cumulative loss to thoroughly disentangle these three factors. 
Our method achieves the state-of-the-art performance on three benchmarks and shows generalization ability in rainy and snowy scenes.
% The source code will be available soon.
\end{abstract}

%%%%%%%%% BODY TEXT

%------------------------------------------------------------------------
\section{Introduction}

\begin{figure}[t]
\centering
\includegraphics[width=\linewidth]{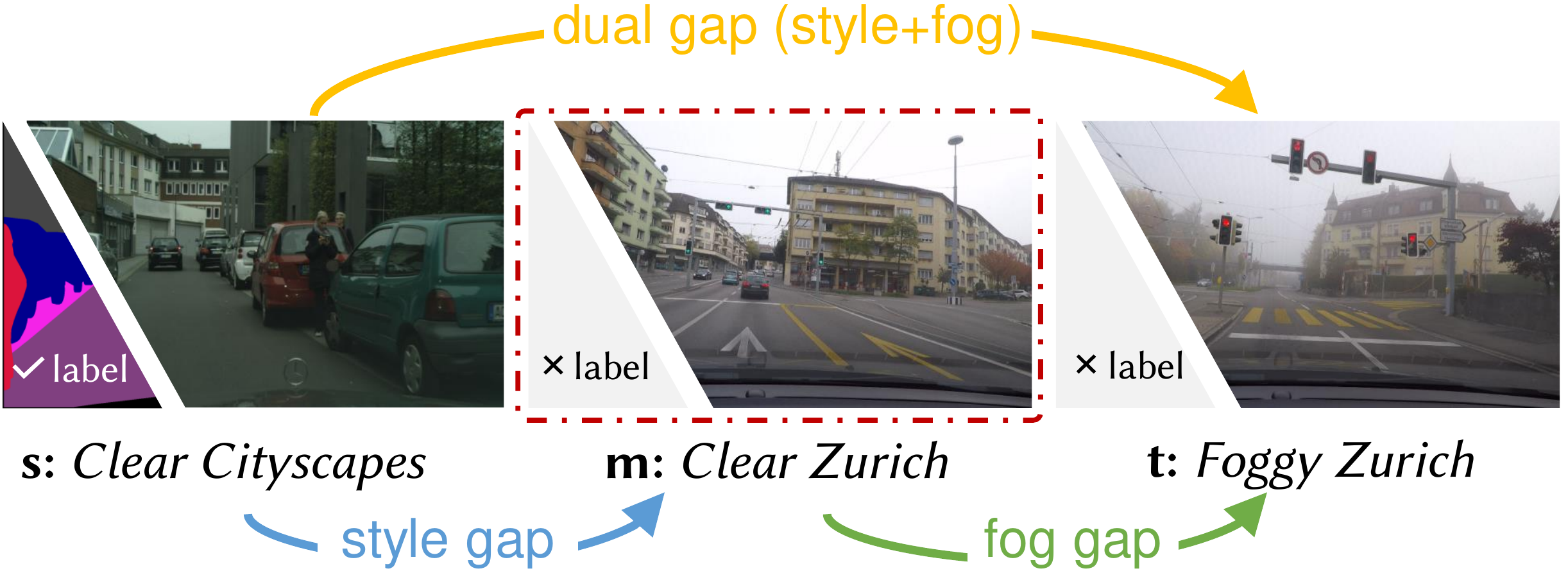}\\
\vspace{-2mm}
\caption{\textbf{The problem and our main idea.} Our goal is to transfer the knowledge from a labeled domain $s$ to an unlabeled domain $\td$. However, direct knowledge transfer is challenging due to the mixed dual-factor gap (orange arrow). By adding an intermediate domain $\imd$ as a bridge, we can decompose the mixed dual-factor gap into two single-factor gaps: the style gap and the fog gap. Since images in both domains $s$ and $\imd$ are captured in clear scenes, we assume there is only the style gap between domains $s$ and $\imd$ (blue arrow). Likewise, images in both domains $\imd$ and $\td$ are collected in the same city (Zurich), we assume there exists only the fog gap between them (green arrow). 
% Based on this, we propose to perform cumulative domain adaptation \ie, narrowing style gap, fog gap and the dual gap one by one. Further, we propose a cumulative relationship constraint to jointly optimize all the steps in a cyclical manner.
}
\vspace{-4mm}
\label{fig:problem}
\end{figure}

% need semantic segmentation --> need robust perceptual --> but foggy effects --> 
% 讲问题很重要
% 难点在哪儿：迁移清晰场景的知识
% Semantic foggy scene understanding (SFSU) is important for outdoor artificial intelligence (AI) applications, such as autonomous driving~\cite{sakaridis2018model,sakaridis2018semantic,dai2020curriculum,2019Purely}.
Semantic foggy scene understanding (SFSU) is important for autonomous driving~\cite{sakaridis2018model,sakaridis2018semantic,dai2020curriculum,2019Purely}.
Although great progress has been made on semantic understanding of clear scenes, SFSU tends to have unsatisfactory performance due to visibility degradation caused by fog~\cite{narasimhan2003contrast,tan2008visibility}.
Besides, unlike the abundant data and annotation under clear scenes, the lack of data and annotation under dense fog weather further complicates this problem.
Therefore, handling the challenging SFSU problem often requires transferring the segmentation knowledge learned from labeled \emph{clear} images to the unlabeled \emph{foggy} images. 

Intuitively, we may tackle this problem by closing the domain gap between clear images and foggy images with state-of-the-art domain adaptation methods.
However, these methods mainly \emph{align} domains in an adversarial~\cite{tsai2018learning,Zhang_2018_CVPR,Chen_2018_CVPR,vu2019advent,chang2019all,Kim_2020_CVPR,li2020content,truong2021bimal,wang2020classes,yang2020fda} or a self-training~\cite{Choi_2019_ICCV,mei2020instance,tranheden2021dacs,araslanov2021self,guo2021metacorrection,gao2021dsp,zhang2021prototypical} manner, 
regardless of how the domain gap is caused. Besides, as has been validated by \cite{sakaridis2018model}, they fail to address the SFSU problem well due to the \emph{large} domain gap.

% ~\cite{tsai2018learning,Zhang_2018_CVPR,Chen_2018_CVPR,Choi_2019_ICCV,Kim_2020_CVPR,li2020content,alshammari2020competitive,vu2019advent,chang2019all,tranheden2021dacs,araslanov2021self,guo2021metacorrection}. 
% An intuitive solution for this problem is to narrow down the gap between clear images and foggy images with the state-of-the-arts domain adaptation methods~\cite{tsai2018learning,Chen_2017_ICCV,Zhang_2018_CVPR,Chen_2018_CVPR,Choi_2019_ICCV,Kim_2020_CVPR}. 
% In this way, the knowledge learned from the labeled \emph{clea}
% Naturally, we may first consider the general domain adaptation method to solve this problem, due to the fact that considerable progresses have been made in this area~\cite{tsai2018learning,Chen_2017_ICCV,Zhang_2018_CVPR,Chen_2018_CVPR,Choi_2019_ICCV,Kim_2020_CVPR}.

% % leaving out the details of how the gap has been caused by specific factors. 
% However, as has been validated by \cite{sakaridis2018model}, 
% % the domain-adversarial approach, such as ~\cite{tsai2018learning}, 
% they cannot address the gap effectively in the SFSU problem due to the large domain gap.

% Recently, more attention have been focused on the fog factor, 
Consequently, the attention of SFSU has been focused on the fog factor, 
which is regarded as the dominating cause of the domain gap in the SFSU problem. One solution is to close this gap by defogging real foggy images, using an existing defogging method~\cite{ren2018deep,de2019learning,zhang2018densely,ren2016single,he2010single,berman2016non,ren2018gated}.
Whereas, the defogging method will also introduce artifacts. They act as noise to hinder the domain adaptation to some extent~\cite{2018Does}. Another solution is to add synthetic fog to clear images and learn with these synthetic foggy images and annotations of clear images in a supervised manner~\cite{sakaridis2018model,sakaridis2018semantic,2019Purely,dai2020curriculum,gong2021analogical}. Nevertheless, these rendered synthetic foggy images, not as real as real foggy ones, could also widen the domain gap between clear and foggy images and yield unsatisfactory performance. Moreover, we argue that these methods over-concerned the fog factor while ignoring other factors, which may affect the domain gap in the SFSU problem.

% Thinking out of the box, we explicitly investigate the domain gap in SFSU and 
% The main reason we argue is that these methods over-concerned the fog factor, while ignoring other factors, which may affect the domain gap in the SFSU problem.
% \zhixiang{Differently, we propose a method 1) to avoid directly treating the total domain gap; 2) without using synthetic foggy data or knowledge of defogging, 
% and consider both style and fog factor. The main idea of our method is to introduce an intermediate domain and use it to... We will show why and how in the following.}

Thinking out of the box, we propose to explicitly investigate the domain gap in SFSU 1) to avoid directly treating the total domain gap; 2) without using synthetic foggy data or knowledge of defogging. We assume that the domain gap is caused by the mixed fog influence and style variation, and both of them are important to SFSU. 
That is, we assume there exist the style-related gap and the fog-related gap in the domain gap of SFSU, and we can decompose the mixed dual-factor gap into these two single-factor gaps by adding an intermediate domain.
% only considering either of them cannot close the dual-factor gap successfully. 
Next, we elaborate on why we can disentangle the style-related gap and what relationship lies between the style-related gap and the fog-related gap in the SFSU problem, using the following empirical finding.

\subsection{Motivation}
\begin{figure}[!tp]
    \centering
    \includegraphics[width=\linewidth]{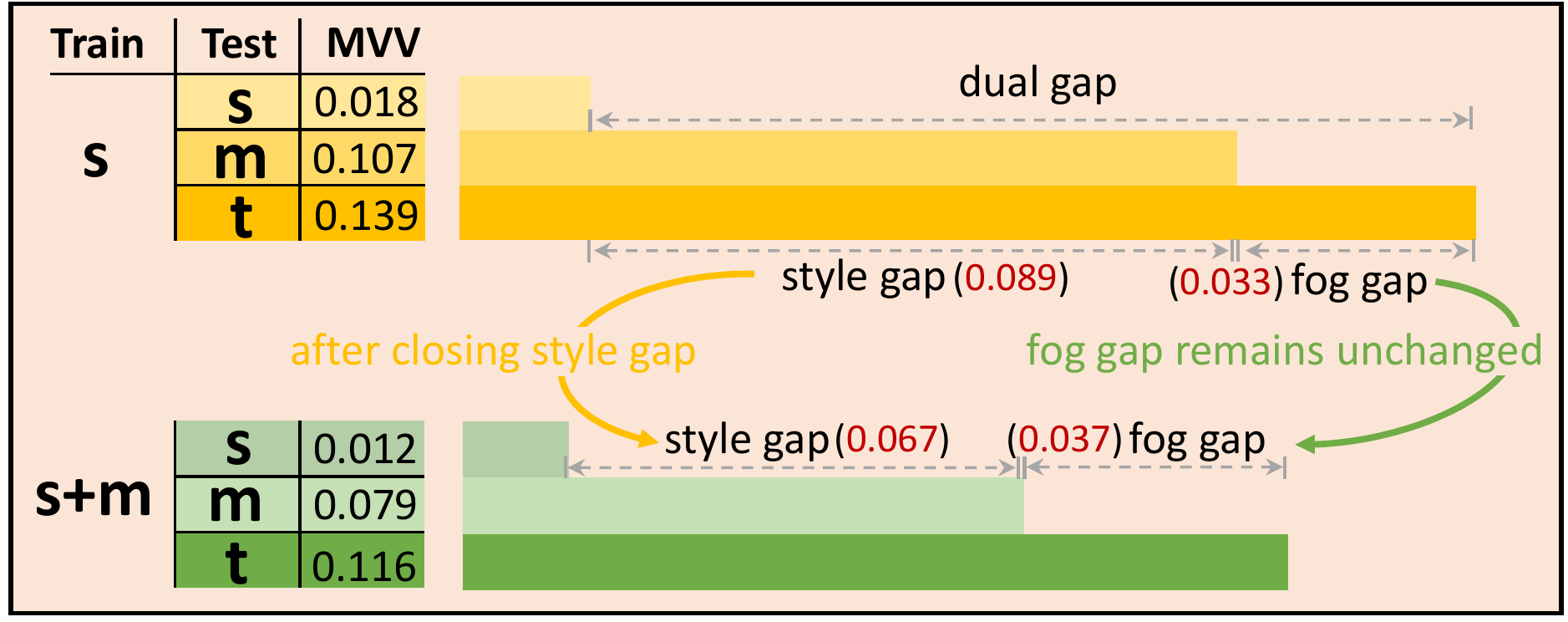}
    \caption{\textbf{Empirical finding of the motivation.} The mean variance value (MVV) measures the overall performance of the segmentation model in a specific domain \ie, domain performance.
    At first, we train a segmentation model with $s$ domain data \ie, this model has learned $s$ domain knowledge. Then, we test it on $s$, $m$ and $t$ domain data, and the performances in three domains are shown as different yellow bars. Besides, the difference between two different bars can represent the performance gap \ie, domain gap (gray dotted arrow), such as the style gap, fog gap and dual gap.
    Next, we adapt the segmentation model with $m$-domain data \ie, this model can learn domain knowledge (related to the style factor) between domains $s$ and $m$, which means the style gap can be closed by this adaptation.
    After this adaptation, the style gap has been closed (from 0.089 to 0.067) while the fog gap remains unchanged (a negligible change of only 0.004). 
    That is, by adding an intermediate domain $\imd$, we disentangle the style gap from the dual gap without damaging the fog gap.
    Thus, we assume that the style gap and fog gap can be divided and closed respectively and the dual gap is an accumulation of the two gaps.}
    \vspace{-4mm}
    \label{fig:empirical}
\end{figure}

% \heading{Motivation.}
We first investigate the influence of the style and fog factors across different domains, \ie, we want to know how the style and fog factors affect the performance of a segmentation model. 
To this end, as shown in Figure~\ref{fig:empirical}, we utilize Mean Variance Value (MVV)\footnote{As has been validated by \cite{zheng2020unsupervised}, the variance, which is calculated from different-level features in a segmentation model, has a strong ability in measuring the uncertainty of a segmentation model when predicting pixel labels. We obtain one Variance Value to represent the uncertainty when the model segments one image. Thus, we calculate the MVV of all images in a specific domain dataset to show the overall performance in this domain.} 
% 脚注说明为什么用MVV来定量表示模型在某个域上的性能
to represent how a segmentation model functions in each domain and how the gap between two different domains is closed. 

Specifically, in Figure~\ref{fig:empirical}, we trained a segmentation model Model ($s$) with $s$-domain data and calculate MVV in domains $s$, $m$ and $t$, yielding $V_{s}^{s}$, $V_{s}^{m}$ and $V_{s}^{t}$, respectively. We use the length of the bar to indicate the performance in each domain.
Ideally, since we only have the model learned from domain $s$, its performance should be good when segmenting images in domain $s$ (\ie, MVV should be low), but tends to degrade when segmenting images in domain $m$ and $t$ (\ie, MVV should be relatively high). Our experiments results are as expected and the yellow bar becomes cumulatively longer when dealing with the images in domain $s$, $m$, and $t$. 
Besides, we can use the difference of two MVVs as the performance gap between two domains (\ie, domain gap). For example, $V_{s}^{m}-V_{s}^{s}$ can represent the gap between domain $s$ and $m$, which we assume as ``style gap''. Likewise, we obtain the ``dual gap'' and ``fog gap''.
% Likewise, we assume $V_{s}^{t}-V_{s}^{s}$ as ``dual gap''. We can easily see that the dual gap is larger than the style gap and the difference between two gaps is around 0.033, which we assume as ``fog gap''.
% We use the length of the bar to indicate how much uncertainty a segmentation model has when it segments image data from individual domain. The longer the bar is, the more uncertainty the segmentation model makes.

Then, we adapt Model ($s$) with $m$-domain data to obtain Model ($s+m$) and calculate the MVV in three domains. Compared with Model ($s$), Model ($s+m$) can learn domain knowledge (related to the style factor) between domains $s$ and $m$ and thus close the style gap (from 0.089 to 0.067). However, the fog gap (0.037) remains large and approximately equals to the fog gap (0.033) before adapting Model ($s$) with $m$-domain data. That is, after closing the style gap, the fog gap still remains unchanged, which means the two gaps can be divided and closed respectively. Meanwhile, the dual gap is always an accumulation of the style gap and the fog gap before and after this adaptation.

Based on this finding, we propose a cumulative domain adaptation framework to address semantic foggy scene understanding, considering both style factor and fog factor in this task. As shown in Figure~\ref{fig:problem}, by adding an intermediate domain $\imd$ as a bridge, we can decompose the mixed dual-factor gap into two single-factor gaps: the style gap and the fog gap. Specifically, we disentangle the style and fog factor separately, and then the dual-factor (style and fog) jointly, which ensures an effective segmentation knowledge transfer from the source domain to the target domain.
Besides, we assume that the dual-factor gap is an accumulation of the style gap and the fog gap. Thus, we further propose a novel cumulative loss to represent this relationship and collaborate the disentanglement of three factors with the cumulative loss in a cyclical manner, enabling our network to transfer segmentation knowledge continuously and further improving the performance.

The main contributions are summarized as follows.
% 分解dual --> decompose  发现问题
% 关系-->
(1) To the best of our knowledge, we are the first to indicate the dual (style and fog) domain gap in SFSU, and also the first to investigate how to close the dual domain gap. Specifically, we propose a novel Cumulative Domain Adaptation (CuDA-Net) method, first disentangling the style factor and fog factor separately, and then the dual-factor jointly. 
% the cumulative relationship of style, fog, and dual factors.  
(2) We find the cumulative relationship of style, fog, and dual factors and thus propose a novel cumulative loss to further disentangle the three factors in a cyclical manner.
(3) Our method outperforms state-of-the-arts on three widely used datasets in SFSU and shows generalization ability on other adverse scenes, such as rainy and snowy scenes.

% (3) We verify the effectiveness of our proposed method on two benchmarks:  \textsc{Foggy-Zurich} and \textsc{Foggy Driving} datasets, yielding +8.4\%, and +12.4\% mIoU improvements over the baseline model, and achieving the state-of-the-art performance.

% We use Figure~\ref{fig:problem} to illustrate our problem setting and the main idea behind the work. As shown in Figure~\ref{fig:problem}, we have images from three datasets, \ie, \textsc{Clear Cityscapes} ($s$), Clear Zurich ($\imd$) and \textsc{Foggy Zurich} ($\td$), which represents domains $s$, $\imd$ and $\td$, respectively. 

% \zhixiang{Elaborate our main idea (what, decompose style and fog), the approach (how, introduce the intermediate domain, note to mention the challenge in method design) and the motivation (why).}

% Hence, from the MVV results in Figure~\ref{fig:empirical}, we can obviously observe that the dual factor is a cumulation of the style factor and the fog factor. The city style and foggy weather both matter in the task of SFSU. 
% Thus, we propose a unified framework to transfer the style and fog simultaneously. Specifically, we disentangle the style and fog factor separately, and then the dual factor (style and fog) jointly. Considering the cumulative relationship revealed above, we collaborate the disentanglement of three factors in a circle to further unleash the potential of the source model. 

\begin{figure*}[t]
\centering
\includegraphics[width=.95\linewidth]{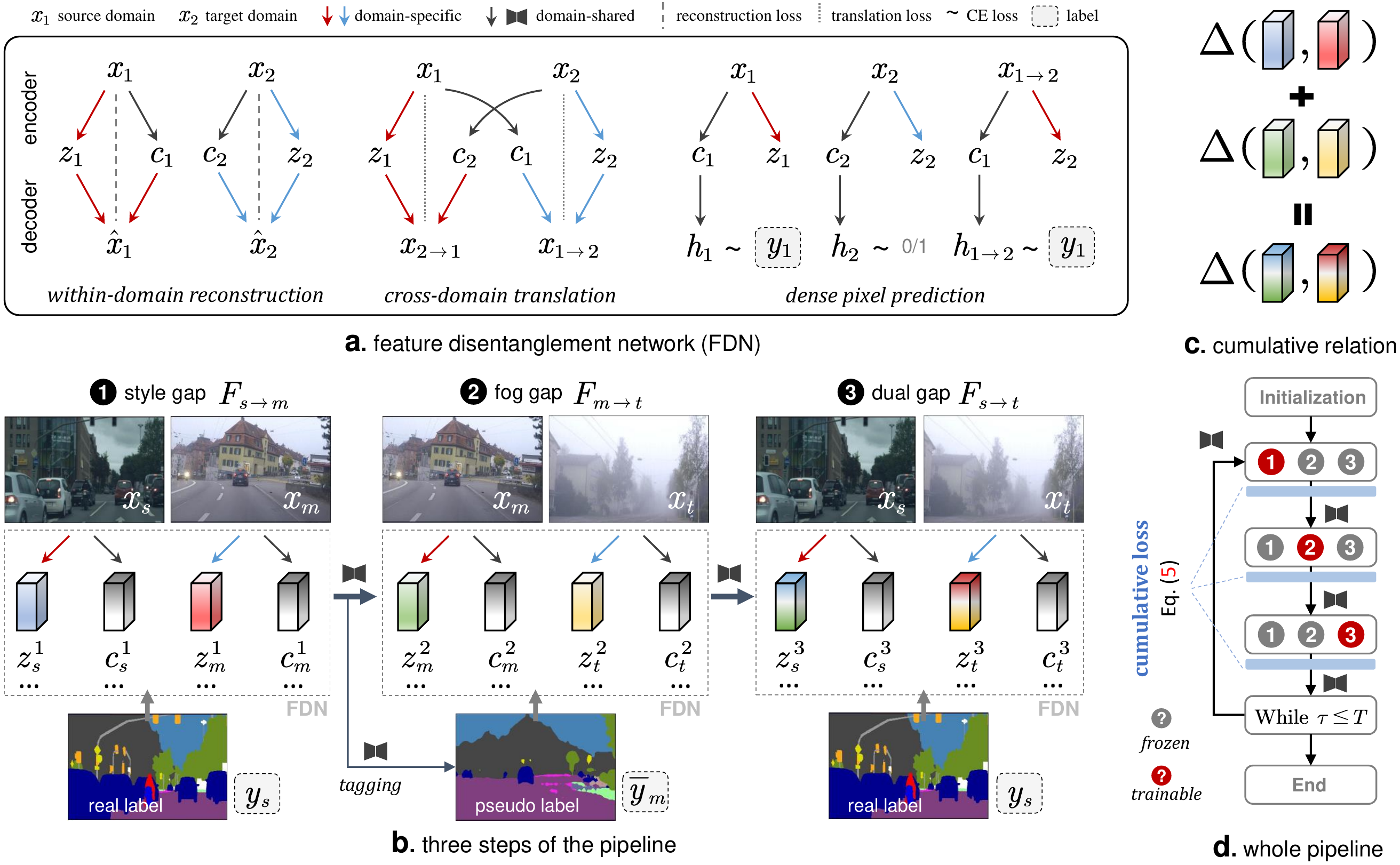}\\
\caption{\textbf{The proposed method}. \textbf{a}, The feature disentanglement network (FDN) disentangles domain-invariant content features from the domain-specific counterparts for images from two different domains. \textbf{b}, By introducing the intermediate domain $\imd$, we can obtain three different input domain combinations, $(x_s, x_\imd)$, $(x_\imd, x_\td)$ and $(x_s, x_\td)$, for three FDNs, $\model_{s\to\imd}$, $\model_{\imd\to\td}$ and $\model_{s\to\td}$, to tackle the style gap, the fog gap and the dual gap respectively.
Three FDNs are trained one by one, where the domain-invariant knowledge is shared.
% to decompose the dual-factor gap into style gap and fog gap. 
% Next, we train  one by one, which are to close the style gap, the fog gap and the dual gap respectively (\textbf{b}). 
As there are no labels for both domain $\imd$ and $t$, we use $\model_{s\to\imd}$ to tag domain $\imd$ for training $\model_{\imd\to\td}$. 
\textbf{d}, The whole pipeline. We first initialize three FDNs by training each of them once, as in \textbf{b}. Then, we conduct cyclical training, using the cumulative relation (\textbf{c}) as an auxiliary loss, for better disentangling the domain-invariant (content) features, which are used to produce segmentation heatmaps.}
\vspace{-2mm}
\label{fig:framework}
\end{figure*}

\section{Method}

% \subsection{Problem Definition}

Suppose that we have $N_s$ labeled images $\{(x_{s}^{i},y_{s}^{i})\}^{N_s}_{i=1}$ from the source domain $s$, 
where $y_{s}^i$ is the label,
% $N_{\imd}$ unannotated intermediate-domain images $\mathcal{X}^{\imd} = \{x^{\imd}_{i}\}^{N_{\imd}}_{i=1}$
and $N_{\td}$ unlabeled images
$\{x_{\td}^{i}\}^{N_{\td}}_{i=1}$ from the target domain $\td$. 
Our goal is to transfer the segmentation knowledge from the source domain $s$ to the target domain $\td$ by our proposed CuDA-Net.
% A key for this problem is 
Motivated by the success of ~\cite{chang2019all}, we use a similar framework as our basic unit to disentangle domain-invariant features from the domain-specific counterparts.
% \textcolor{red}{We follow a previous successful practice that disentangles domain-invariant features from the domain-specific counterparts~\cite{chang2019all}}.
However, since images in domain $s$ and $\td$ are taken under different cities and weathers, they encounter large domain gaps caused by mixed style and fog factors, which challenges this method.
% We propose to learn domain-invariant features by disentangling the style and fog factors.
Therefore, we propose to decompose the mixed factors into separate ones by introducing an intermediate domain $\imd$ with $N_{\imd}$ unlabeled images $\{x_{\imd}^{i}\}^{N_{\imd}}_{i=1}$, which share similar fog influence (no fog) with the source domain and similar style variation with the target domain (same city).
% Pairing them with images from the source domain and the target domain and feeding them into the disentanglement network
Figure~\ref{fig:framework} depicts the framework of our proposed method. 
It includes three sub-networks: 
$\model_{s \rightarrow \imd}$, 
$\model_{\imd \rightarrow \td}$ 
and $\model_{s \rightarrow \td}$, 
which share the same prototypes to disentangle the domain-invariant features from the domain-specific counterparts (Figure~\ref{fig:framework}\textbf{a}).
% But they have different inputs,
They are fed with different input pairs $(x_s, x_\imd)$, $(x_\imd, x_\td)$ and $(x_s, x_\td)$ to close the style gap, the fog gap and the dual gap respectively.
We train them one by one (Figure~\ref{fig:framework}\textbf{b}) and share the domain-invariant knowledge forward. After training these three sub-networks (initialization), we conduct a cyclical training (Figure~\ref{fig:framework}\textbf{d}) using the cumulative relation (Figure~\ref{fig:framework}\textbf{c}) as auxiliary loss, to help better disentangle the domain-invariant (content) features, which are used to produce segmentation heatmaps.
% and other three domain specific features(style, fog and dual).

% Domain-specific and domain-invariant 
\subsection{Feature Disentanglement Networks}
\label{sec:FDN}

Feature Disentanglement Networks (FDN) is the basic unit of our method, as shown in Figure~\ref{fig:framework}\textbf{a}. 
Given images $x_1$ and $x_2$ from two different domains, with the ``shared content space'' assumption \cite{huang2018multimodal}, it can disentangle domain-invariant content features $c_1$ and $c_2$ of these images from the domain-specific counterparts $z_1$ and $z_2$. As has been validated by \cite{chang2019all}, the content features contributes most to the semantic segmentation task. Therefore, through feature disentanglement, we can transfer the segmentation knowledge from $x_1$ domain to $x_2$ domain. 

Specifically, we first use a shared content encoder $E_{c}$ (black line) to extract $c_1$ and $c_2$ and two private encoders to extract domain-specific feature $z_1$ and $z_2$ respectively (red and blue line). Then, we use an shared image decoder $D$ to decode an image using the content features $c_{1}$, $c_{2}$, and domain-specific feature $z_{1}$, $z_{2}$. Depending on which $c$ and $z$ we use, we can perform within-domain reconstruction and cross-domain translation to supervise the disentanglement learning. Besides, we use a segmentation head $S$ to produce segmentation heatmaps $h$ from the content feature $c$, where label $y_{1}$ is used as the supervision signal.

% \textcolor{blue}{
We build our FDN with a similar framework as DISE~\cite{chang2019all} because  both of us adopt the ``shared content space'' assumption~\cite{huang2018multimodal}. However, we only design four necessary losses to train our FDN, aiming to enable the FDN to close three different gaps (style gap, fog gap and dual gap). While, DISE~\cite{chang2019all} utilizes seven losses to close one gap between synthetic clear data and real clear data, which is time-consuming to train and hard to converge.
% }

\heading{Within-domain reconstruction.} 
% As shown in the left of Figure~\ref{fig:framework}\textbf{a}, 
% If $c$ and $z$ are from the same domain image,
% We expect content feature $c$ and private feature $z$ extracted from each domain image can be used to perfectly reconstruct the original one. 
We expect images decoded using content feature $c$ and private feature $z$ extracted from the same image can perfectly reconstruct the original ones. 
Thus, we define the reconstruction loss as:
\vspace{-1mm}
\begin{equation} 
{L}_{rec} =  {L}_{pixel}(x_{1}, \hat{x}_1) + {L}_{pixel}(x_{2}, \hat{x}_2)\,,
\vspace{-1mm}
\end{equation}
where the pixel-wise loss ${L}_{pixel}(\cdot, \cdot)$ is implemented by the perceptual loss~\cite{simonyan2014very} with shadow layer features highlighted. 
% It also can be replaced with other pixel-wise losses, e.g., $L_2$.

\heading{Cross-domain translation.} 
% As is shown in the middle of Figure~\ref{fig:framework}\textbf{a},
We recombine the content feature $c$ from one domain image and private feature $z$ from another domain image to generate the translated image.
% which is called cross-domain translation.
For example, in our sub-network $\model_{\imd \rightarrow \td}$ in Figure~\ref{fig:framework}\textbf{b}, by recombining the content feature $c_t^2$ from $x_\td$ and the private feature $z_m^2$ from $x_\imd$, we can generate an image, which can be regarded as the defogged version of $x_\td$.
% Likewise, we can synthesize fog on $x_s$.
For the translated images ${x}_{1 \rightarrow 2}$ and ${x}_{2 \rightarrow 1}$ whose private features have been changed, we impose content consistency losses $L_{con}$, which is implemented by the perceptual loss ~\cite{simonyan2014very} with the deep layer features highlighted, to constrain the content aspect of translated images and original images:
\vspace{-1mm}
\begin{equation} 
{L}_{trans} =  {L}_{con}(x_{1}, {x}_{1 \rightarrow 2}) + {L}_{con}(x_{2}, {x}_{2 \rightarrow 1})\,.
% \vspace{-1mm}
\end{equation}
% In addition, we follow LSGAN~\cite{mao2017least} and Patch Discriminator~\cite{isola2017image} using an adversarial loss ${L}_{transadv}$ to guarantee the translated images look as if they were from the domain where its content belongs. 
% We follow LSGAN~\cite{mao2017least} and Patch Discriminator~\cite{isola2017image} to achieve this.
% \zhixiang{TODO: to details the within-domain reconstruction and cross domain translation....(refer to MUNIT or other image2image translations for writing)
% Follwing DISE, we use DIRT? It can be replaced with other image-to-image translation methods.}

\heading{Dense pixel prediction.}
Thanks to the domain invariance discovered by disentanglement learning, we can transfer the semantic knowledge across domains.
% Since the domain-invariant features, $c_1$ and $c_2$, , 
We apply the segmentation head $S$ on $c_1$ and $c_2$ to obtain the probability outputs of each pixel $h_1, h_2\!\in\! \mathbb{R}^{H\times W \times C}$, 
% and $h_2 \!\in\! \mathbb{R}^{H\times W \times C}$, 
where $H, W, C$ represents the height, the width and the number of class categories, respectively.
To supervise the training of the shared content encoder $E_{c}$ and the segmentation head $S$,
we use the cross-entropy to calculate segmentation loss ${L}_{seg}^1$ between $h_1$ and its corresponding label $y_1$.
% is utilized to measure the distance between the source domain segmentation probability $h_1$ and its corresponding ground-truth label $y_1$ with the cross-entropy:
% is given by the cross-entropy based on ground truths in the domain ``1''
% \begin{equation} 
% %  \begin{split}
% {L}_{seg}^1 =  -\sum_{h, w} \sum_{c \in C} (y_1)^{(h, w, c)} \log ((h_1)^{(h, w, c)})\,.
% % \end{split}
% \vspace{-2mm}
% \end{equation}
Besides, since the $1$-domain-like image ${x}_{1 \rightarrow 2}$ share the same content as $x_{1}$, the labels $y_1$ can be the pseudo labels for ${x}_{1 \rightarrow 2}$. Hence, we calculate ${L}_{seg}^{1 \rightarrow 2}$ between $h_{1 \rightarrow 2}$ and label $y_1$, also using the cross-entropy.
% \begin{equation} 
% {L}_{seg}^{1 \rightarrow 2} = -\sum_{h, w} \sum_{c \in C} (y_1)^{(h, w, c)} \log ((h_{1 \rightarrow 2})^{(h, w, c)})\,.
% % \vspace{-2mm}
% \end{equation}

Aside from supervised losses, an adversarial loss ${L}_{segadv}$ at the output of the segmentation head $S$ is introduced, in the hopes of making the content encoder $E_{c}$ and the $S$ generalize well on the domain $2$. To this end, we introduce the domain discriminator $\mathit{Dis}$ and fool $\mathit{Dis}$ by maximizing the probability of target domain prediction $h_{2}$ being considered as the source domain prediction:
%The details refer to \cite{chang2019all}.
% That is,
\begin{equation}
{L}_{segadv}=-\sum_{h, w} \log \left(\mathit{Dis}\left(h_{2}\right)^{(h, w, 1)}\right)\,,
\end{equation}
where $1$ means the discriminator $\mathit{Dis}$ perceive $h_{2}$ as the source domain prediction. 
% \textcolor{red}{The detailed structure of discriminator $\mathit{Dis}$ and content encoder $E_{c}$ is shown in supplementary material.}
% We use the DeepLab-v2 as our segmentation head. However, we note that our method can easily integrate other architectures as well.

\heading{Feature disentanglement loss.}
The disentanglement loss function in FDN is a weighted combination of each loss:
% \begin{equation} 
% \begin{split}
% {L}_{m \rightarrow n} =& \lambda_{rec}{L}_{rec} + \lambda_{trans}{L}_{trans} + \lambda_{seg}^{1}{L}_{seg}^{1} \\
% +& \lambda_{segadv}{L}_{segadv} + \lambda_{seg}^{1 \rightarrow 2}{L}_{seg}^{1 \rightarrow 2},\\
% \end{split}
% \label{equa:full_loss}
% \end{equation}
\begin{equation} 
\begin{split}
{L}_{1 \rightarrow 2} =& \lambda_{rec}{L}_{rec} + \lambda_{trans}{L}_{trans} \\
+& \lambda_{seg}({L}_{seg}^{1}+{L}_{seg}^{1 \rightarrow 2}) 
+ \lambda_{segadv}{L}_{segadv}
\end{split}\,,
\label{equa:stage1_loss}
\end{equation}
where ${L}_{1 \rightarrow 2}$ can be ${L}_{s\rightarrow\imd}$, ${L}_{\imd \rightarrow\td}$ or ${L}_{s\rightarrow\td}$ for the following disentanglement and the weights $\lambda_{rec}$, $\lambda_{trans}$, $\lambda_{seg}$ and $\lambda_{segadv}$ are empirically set as $0.5$, $0.1$, $1$ and $1$ to control the relative importance of reconstruction/translation quality, the prediction accuracy and domain generalization.

\subsection{Style and Fog Decomposition}
% 为啥要分style和fog: 直接close dual不行 假设中间域
% 先讲FDN怎么用到style上，同理到fog 和 dual上
% 再讲1->2的知识传递: shared content encoder、decoder、segmentation head，Discriminator 唯一差别就是 得到各个域的private encoder
% 额外一点：step2的伪标签是根据step1 trained content encoder 和 segmentation head得到的
% 2->3: 同1到2的知识传递  done

% 3->4: freeze 其他两个域的private encoder、 decoder；update 其他 ；额外引入加和关系
% 迭代：10K T=2

% 
% 要提中间域是如何划分的 人为选的 从目标域中划分出了一部分 网络也选同等数目图片
% 多域工作的调研

%\heading{Step $S\to\imd$ (Style)}

The aforementioned FDN is designed to transfer the segmentation knowledge by disentangling the domain-invariant features and domain-specific features. 
However, directly applying FDN to domain $s$ and domain $t$ cannot achieve ideal performance.
The reason we suppose is that the mixed dual-factor gap between domain $s$ and domain $t$ is too large to close, which is also the weakness of other domain adaptation methods. 
% which we assume the mixed dual-factor gap between domain $s$ and domain $t$ is too large to close by FDN. 
Thus, we introduce the intermediate domain $m$, decomposing the dual-factor gap into two single-factor gaps: style and fog. Since images in both domain $s$ and $\imd$ are captured in clear scenes, we assume there is only the style gap between them. Similarly, as images in both domain $\imd$ and $\td$ are collected in the same city (Zurich), we assume there exists only the fog gap between them. Therefore, we use three sub-network $\model_{s \rightarrow \imd}$, $\model_{\imd \rightarrow \td}$ and $\model_{s \rightarrow \td}$ to disentangle the style factor, fog factor and dual-factor one by one and gradually transfer the segmentation knowledge from domain $s$ to $\td$ (Figure~\ref{fig:framework}\textbf{b}).

Concretely, $\model_{s \rightarrow \imd}$ first utilizes two specific private style encoders $E_{\sty}^{s}$ and $E_{\sty}^{\imd}$ to extract the latent style features $z_{s}^{1}$ and $z_{m}^{1}$, respectively. 
The labels $\{y_{s}^i\}_{i=1}^{N_s}$ supervise the training process.
% After training with domain $s$ and domain $\imd$ images and , 
After then, 
except for the two private style encoders, the remaining part of this trained $\model_{s \rightarrow \imd}$, which we perceive as domain shared part and represents segmentation knowledge, will be passed to the next sub-network $\model_{\imd \rightarrow \td}$. In other words, an content encoder $E_{c}$, segmentation head $S$, image decoder $D$ and domain discriminator $\mathit{Dis}$ are used as initialization of sub-network $\model_{\imd \rightarrow \td}$. Note that the domain $\imd$ has no labels, we used the trained $\model_{s \rightarrow \imd}$ to generate pseudo labels for training the $\model_{\imd \rightarrow \td}$. After training $\model_{\imd \rightarrow \td}$, except for the two fog encoders $E_{\fog}^{\imd}$ and $E_{\fog}^{\td}$, the domain shared part of  $\model_{\imd \rightarrow \td}$ are used as the initialization of sub-network $\model_{s \rightarrow \td}$. Likewise,  $\model_{s \rightarrow \td}$ uses two dual-factor (style and fog) encoders $E_{\dual}^{s}$ and $E_{\dual}^{\td}$ to extract the latent dual-factor features $z_{s}^{3}$ and $z_{t}^{3}$, respectively. 
To put it simply, through the training of $\model_{s \rightarrow \imd}$, $\model_{\imd \rightarrow \td}$ and $\model_{s \rightarrow \td}$, we pass down the segmentation knowledge from domain $s$ to domain $\td$ in a more effective way and obtain three pairs of domain-specific feature encoders for further feature disentanglement in the cumulative domain adaptation.

% \includesvg[scale=0.35]{img/logo/zm1.svg}   \includesvg[scale=0.35]{img/logo/zm2.svg} \includesvg[scale=0.35]{img/logo/zt2.svg} \includesvg[scale=0.35]{img/logo/zt3.svg}
% \includesvg[scale=0.35]{img/logo/zs1.svg} \includesvg[scale=0.35]{img/logo/zs3.svg}
% \includesvg[scale=0.35]{img/logo/content.svg}

% \subsection{Loss for Accumulation Relationship Constraint}
\subsection{Cumulative Domain Adaptation}

% \zhixiang{Why Cumulative Domain Adaptation}

\heading{Cumulative loss.}
As verified in our motivation, 
there exists a cumulative relationship among three kinds of domain factors (private features). As shown in Figure~\ref{fig:framework}\textbf{c}, if we take $\Delta(z_{\imd}, z_{s})$ as the style discrepancy between domain $\imd$ and $s$, take $\Delta(z_{\td}, z_{\imd})$ as the fog discrepancy between domain $\td$ and $\imd$, and take $\Delta(z_{\td}, z_{s})$ as the dual discrepancy between domain $\td$ and $s$, it is reasonable to assume that the dual discrepancy is a cumulation of the style and fog discrepancies, namely, 
$\Delta(z_{\imd}, z_{s})+\Delta(z_{\td}, z_{\imd}) = \Delta(z_{\td}, z_{s})$. 
Thus, we design the cumulative relationship loss function as:
\begin{equation}
\label{eq:cumulation}
\begin{split}
{L}_{cum} = \left \| \Delta(z_{\imd}, z_{s}) + \Delta(z_{\td}, z_{\imd}) - \Delta(z_{\td}, z_{s}) \right \| ^2\,.
\end{split}
\vspace{-3mm}
\end{equation}
% We also tried different penalties, e.g., $L_1$. 
Then, we take a step further and use this cumulative loss ${L}_{cum}$ as an additional loss to conduct our proposed cumulative domain adaptation, by utilizing private encoders trained in the first three steps before.

% 3->4: freeze 其他两个域的private encoder、 自身的decoder；update 其他 ；额外引入加和关系约束 
% 迭代：10K T=2
% 循环增强 三对private encoders, 进而进一步增强content encoder

% \begin{figure}
%     \centering
%     \includegraphics[width=\linewidth]{img/accumu_v4.pdf}
%     \caption{The cumulative training pipeline.}
%     \label{fig:accumu}
% \end{figure}

% \subsection{Training Pipeline}
\heading{Training pipeline.}
Figure~\ref{fig:framework}\textbf{d} depicts the whole training process.
The three trained sub-network $\model_{s \rightarrow \imd}$, $\model_{\imd \rightarrow \td}$ and $\model_{s \rightarrow \td}$ are used as the initialization of the cumulative domain adaptation. Specifically, we use all modules in $\model_{s \rightarrow \td}$ (an content encoder, two dual-factor encoders, an image decoder and an segmentation head), two style encoders in $\model_{s \rightarrow \imd}$ and two fog encoders in $\model_{\imd \rightarrow \td}$ to build up the whole network. Next, as shown in Figure~\ref{fig:framework}\textbf{b}, we input $(x_s, x_\imd, x_\td)$ tuples for extracting style, fog and dual private features. Then, we freeze the two fog encoders and two dual-factor encoders and train other modules (especially two style encoders and content encoders) in the whole network, using the final loss below: 
\vspace{-2mm}
\begin{equation} 
\vspace{-2mm}
\begin{split}
{L}_{final} =  {L}_{s \rightarrow \imd} +  \lambda_{cum}{L}_{cum} \,.
\end{split}
\label{equa:stage2_loss}
\vspace{-4mm}
\end{equation}
After this training, we assume the style encoders can capture domain-specific style features better due to combining feature disentanglement loss ${L}_{s \rightarrow \imd}$ with the cumulative loss ${L}_{cum}$ and the content encoders can better extract shared content features, which is used to produce the segmentation heatmap. 
The following two steps in Figure~\ref{fig:framework}\textbf{d} have the same function and the only difference is which private encoders we train and which encoders we freeze. Note that the shared content encoder is always trainable in the three steps, and we use the content encoder to update pseudo labels for training fog encoders. Besides, we train the whole network in a cyclical manner, hoping to improve the disentangling ability of three pairs of private encoders alternatively and continuously enhance the shared content encoder. Empirically, we set $T$ as 3, which means we conduct the cyclical cumulative training three times. Finally, we use the trained content encoder and segmentation head $S$ in $\model_{s \rightarrow \td}$ to produce the segmentation heatmaps for testing.

\section{Experiments}
\subsection{Datasets}
% \begin{itemize}[leftmargin=*]
\noindent\textbf{Cityscapes}~\cite{marius2016cityscapes} is a real-world dataset composed of street view images captured in 50 different cities. Its data split includes 2975 training images and 500 validation images.
% with each having a spatial resolution of $2048\!\times\!1024$ and 19 semantic labels at the pixel level. 
    % Following selection criteria for better fog simulation~\cite{sakaridis2018semantic}, we choose 498 clear images from the training set as the source domain dataset, named as Clear Cityscapes}. Note that images in Clear Cityscapes are not captured in Zurich city.
    
\noindent\textbf{Foggy Cityscapes DBF}~\cite{sakaridis2018semantic} has 550 synthetic foggy images in total, including 498 training images and 52 testing images. These images are selected from Cityscapes and synthesized with fog using depth information. We use 498 clear images from Cityscapes as the source domain dataset, named as \textbf{Clear Cityscapes}. Note that images in Clear Cityscapes are not captured in Zurich city.
    
\noindent\textbf{Foggy Zurich*}~\cite{sakaridis2018model} contains 3808 real-world foggy road scenes in the city of Zurich and its suburbs. 
% The spatial resolution of each image is $1920\!\times\!1080$.
According to fog density, it is split into two categories --- light and medium, consisting of 1552 images and 1498 images, respectively. We use the medium category as the target domain dataset, named as \textbf{Foggy Zurich}. Besides, it has a test set--Foggy Zurich-test including 40 images with labels that are compatible with Cityscapes~\cite{marius2016cityscapes}.

\noindent\textbf{Foggy Driving}~\cite{sakaridis2018model}. It is a collection of 101 real-world foggy road-scenes images, in which 33 images are finely annotated and the rest 68 images are coarsely annotated. They are purely used for testing. 
    
\noindent\textbf{Clear Zurich}. We manually select 248 images from the light category of Foggy Zurich*~\cite{sakaridis2018model} and term this dataset as Clear Zurich. We use this \textbf{Clear Zurich} as an intermediate domain dataset because we perceive these images visually as clear scene images.
 
\noindent\textbf{ACDC}~\cite{ACDC}. It contains four adverse-condition categories (fog, rain, snow and nighttime) with pixel-level annotations. Each of them contains 1000 images and is split into train set, validation set and test set for roughly 4:1:5 proportion. The test set is withheld for testing online. 

\begin{table}[t]
\caption{\textbf{Performance comparison}. Experiments are conducted on Foggy Zurich (FZ) and Foggy Driving (FD), measured with mean IoU (mIoU \%)  over all classes. For results on ACDC, please refer to the ACDC-fog benchmark website.
% Note that these methods do not use synthetic foggy images at all. And our method CuDA-Net obtain the top performance using the same training data.
% {$^\dagger$Since the synthesis-based methods use additional synthetic foggy images, for fair comparison, we also used these images like CMAda3+, named as CuDA-Net+. 
% But it is worth noting that our CuDA-Net+ outperforms CMAda3+ only using one third of synthetic foggy images CMAda3+ used.}
}
\label{tab:sota_1}
\vspace{-6mm}
\begin{center}
\renewcommand\tabcolsep{6pt}
\resizebox{\linewidth}{!}{
\begin{tabular}{lcl|cc}
\toprule
Experiment    & Method & Backbone & FZ & FD\\
\midrule
\multirow{2}*{\emph{Backbone}} 
            & -- & DeepLab-v2~\cite{lin2017refinenet} &  25.9        &   35.7   \\
           & --  & RefineNet~\cite{chen2017deeplab} &   34.6        &  35.8     \\
           \midrule
\multirow{5}*{\emph{Defogging}} & MSCNN~\cite{ren2016single}     & RefineNet  &34.4  & 38.3 \\
           & DCP~\cite{he2010single}       & RefineNet  &31.2  & 33.2 \\
           & Non-local~\cite{berman2016non} & RefineNet  &27.6  & 32.8 \\
           & GFN~\cite{ren2018gated}       & DeepLab-v2 & 27.5     & 37.2      \\
           & DCPDN~\cite{zhang2018densely}  & DeepLab-v2 & 28.7     & 37.9     \\
           \midrule 
\multirow{9}*{\shortstack[l]{\emph{Domain}\\ \emph{Adaptation}}} &                 Multi-task~\cite{alshammari2020competitive} &     --       &   26.1        &  31.6      \\
           & AdSegNet~\cite{tsai2018learning}   & DeepLab-v2 &   26.1        &  37.6     \\
           & ADVENT~\cite{vu2019advent}     & DeepLab-v2 &   24.5       &  36.1        \\
           & DISE~\cite{chang2019all}       & DeepLab-v2 &   40.7        &  45.2    \\
           & CCM~\cite{li2020content}       & DeepLab-v2 &   35.8        &  42.6    \\
           & SAC~\cite{araslanov2021self} & DeepLab-v2      &   37.0        &  43.4    \\
           & ProDA~\cite{zhang2021prototypical} & DeepLab-v2    &   37.8        &  41.2    \\
           & DMLC~\cite{guo2021metacorrection} & DeepLab-v2 &   33.5        &  32.6    \\
           & DACS~\cite{tranheden2021dacs} & DeepLab-v2 &   28.7        &  35.0    \\
           \midrule
\emph{Defogging+DA} & MSCNN~\cite{ren2016single}+DISE~\cite{chang2019all} & DeepLab-v2 &   38.6 &  37.1    \\ 
            \midrule
\rowcolor{Gray} \emph{Ours}       & CuDA-Net & DeepLab-v2 & \textbf{48.2}    &  \textbf{52.7}   \\
          \midrule
\multirow{6}*{\emph{Synthesis}$^\dagger$}   & SFSU~\cite{sakaridis2018semantic}  & RefineNet &   35.7   &  35.9     \\
          & CMAda2~\cite{sakaridis2018model}    & RefineNet &   42.9        &  37.3     \\
          & CycleGAN~\cite{zhu2017unpaired}  & RefineNet &   40.5   &  47.7  \\
          & MUNIT~\cite{huang2018multimodal}     & RefineNet &   39.1   &  47.8  \\
          & AnalogicalGAN~\cite{gong2021analogical} & RefineNet & 42.3   &  47.5  \\
          & CMAda3+~\cite{dai2020curriculum}   & RefineNet &   46.8        &  49.8     \\
          \midrule
\emph{Synthesis+DA} & SFSU~\cite{sakaridis2018semantic}+DISE~\cite{chang2019all} & DeepLab-v2 &   39.3 &  39.0    \\ 
\midrule
\rowcolor{Gray}
\emph{Ours}      & CuDA-Net+ & DeepLab-v2 &  \textbf{49.1}    &  \textbf{53.5}   \\
\bottomrule
\end{tabular}}
\end{center}
\vspace{-3mm}
{\raggedright \footnotesize{$^\dagger$Since the synthesis-based methods use additional synthetic data, for fair comparison, we also add these data to train our sub-network $\model_{m \rightarrow \td}$ before cumulative domain adaptation, named CuDA-Net+.} \par}
\vspace{-2mm}
\end{table}

\begin{table}[t]
\begin{center}
\renewcommand\tabcolsep{1pt}
\caption{\textbf{Training data comparison with CMAda3+.} Both our CuDA-Net and CuDA-Net+ outperform CMAda3+, using less synthetic foggy data and less real foggy data. `light', `medium' and `dense' in the table indicates the different fog density.}
\label{tab:images_to_use}
\vspace{-3mm}
\resizebox{\linewidth}{!}{
\begin{tabular}{lccc}
\toprule
Training data used        & CMAda3+    & CuDA-Net   & CuDA-Net+   \\
\midrule
Clear Cityscapes    & 498        & 498           & 498         \\
\midrule
\multirow{3}{*}{\makecell[l]{Foggy Cityscapes DBF\\(synthetic fog)}} &  498 (light) & --  & -- \\
& 498 (medium) & -- & -- \\ 
& 498 (dense) & -- & 498 (dense) \\
\midrule
\multirow{2}{*}{\makecell[l]{Foggy Zurich*\\(real fog)}}        &1552 (light) & 248 (light) & 248 (light)\\
& 1498 (medium)  &  1498 (medium) &  1498 (medium) \\
\midrule
Total Number   & 5042 & 2244 & 2742 \\
\midrule
mIoU (on FZ)  & \textbf{46.8} & \textbf{48.2} & \textbf{49.1} \\
\bottomrule
\end{tabular}}
\vspace{-8mm}
\end{center}
\end{table}

\begin{figure*}[t]
\centering
		\tabcolsep=0.5pt
		\renewcommand\arraystretch{0.5}
		\begin{tabular}{cccccc}
			\includegraphics[width=0.16\textwidth]{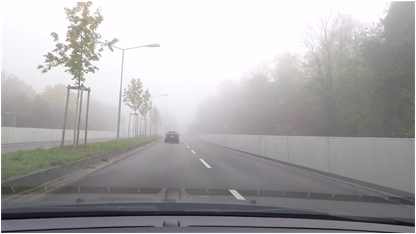} &
			\includegraphics[width=0.16\textwidth]{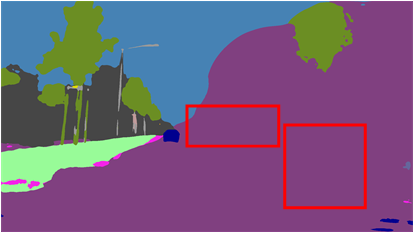} &
			\includegraphics[width=0.16\textwidth]{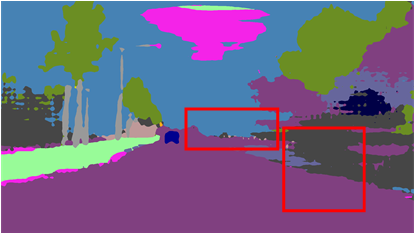} &
			\includegraphics[width=0.16\textwidth]{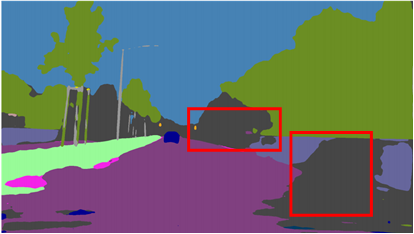} &
			\includegraphics[width=0.16\textwidth]{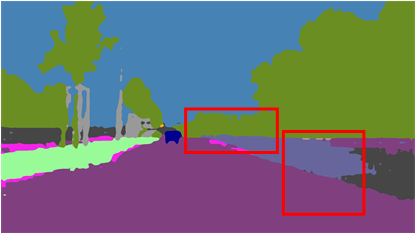} &
			\includegraphics[width=0.16\textwidth]{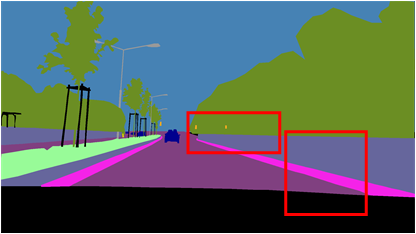} \\
			\includegraphics[width=0.16\textwidth]{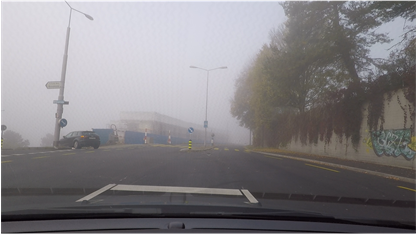} &
			\includegraphics[width=0.16\textwidth]{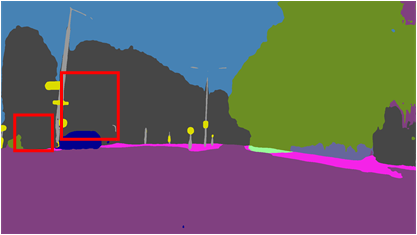} &
			\includegraphics[width=0.16\textwidth]{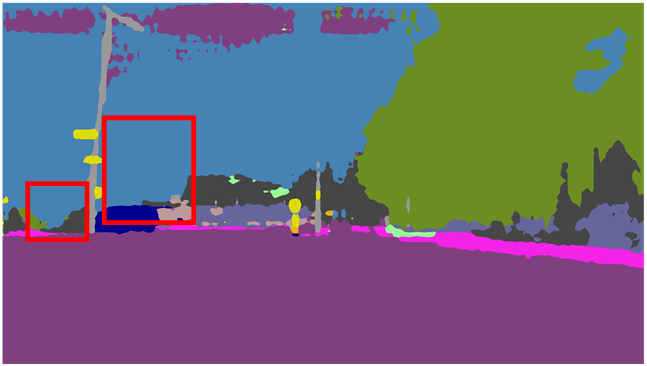} &
			\includegraphics[width=0.16\textwidth]{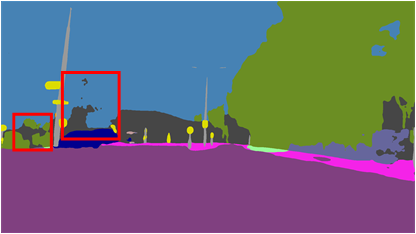} &
			\includegraphics[width=0.16\textwidth]{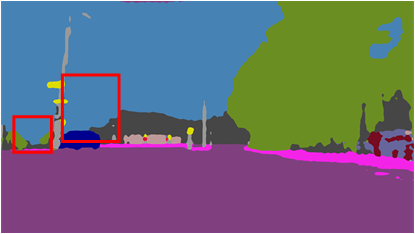} &
			\includegraphics[width=0.16\textwidth]{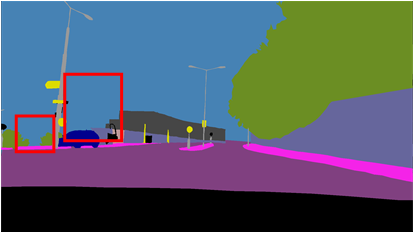} \\
			Input  & MSCNN & DISE & CMAda3+ & CuDA-Net (ours) & Ground Truth   \\
	\end{tabular}
	\vspace{-2mm}
\caption{\label{fig:sota} \textbf{The qualitative comparison with the SOTA methods}. The input images are randomly selected from Foggy Zurich-test. The red boxes clearly show that our method can better deal with the details than the SOTA methods.} 
\vspace{-2mm}
\end{figure*}

\begin{figure}[!t]
\centering
\includegraphics[width=\linewidth]{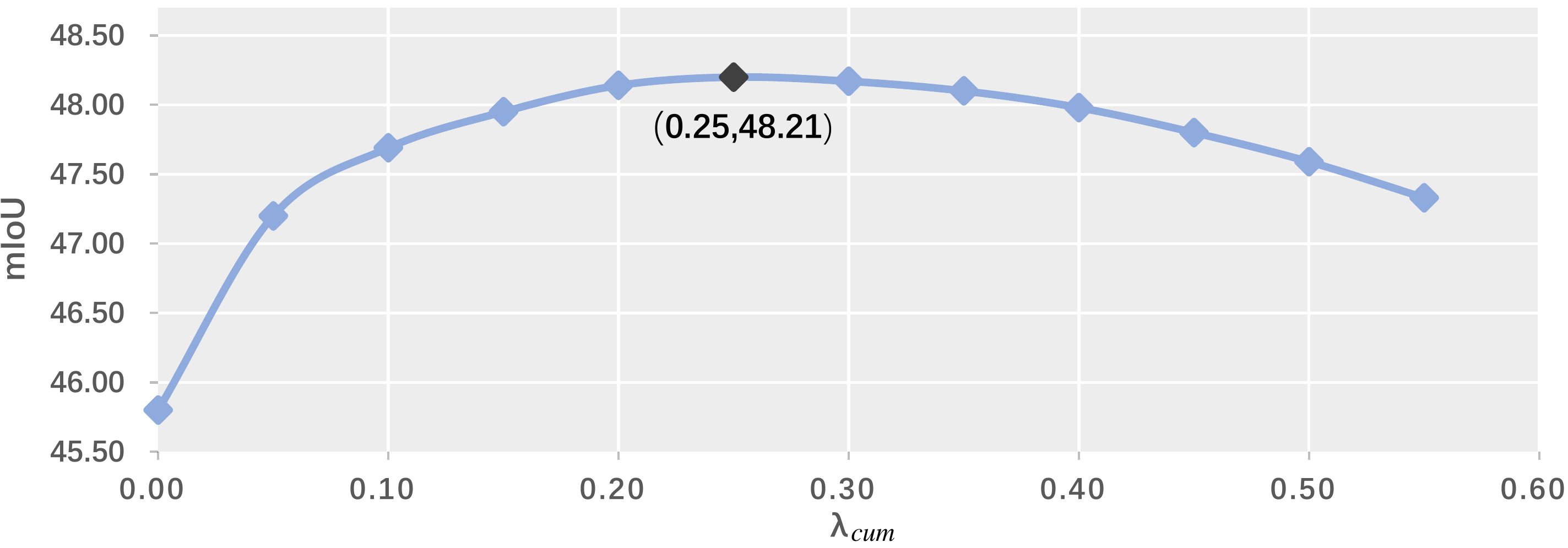}
\vspace{-4mm}
\caption{\textbf{Ablation study on $\lambda_{cum}$ in Eq.~(\ref{equa:stage2_loss})} on Foggy Zurich-test dataset. The results show our model is not sensitive to $\lambda_{cum}$.}
\vspace{-2mm}
\label{fig:lamda}
\end{figure}

\begin{table}[!t]
    \caption{\textbf{Ablation study}. These experiments are performed on Foggy Zurich-test dataset.}
	\label{tab:ablation}
	\vspace{-2mm}
	\centering
	\resizebox{1\linewidth}{!}{
		\begin{tabular}{l|ccc|cc}
		\toprule
		              &  \multicolumn{3}{c|}{Components} & mIoU & gain \\  \midrule
		Initialization & \multicolumn{3}{c|}{Deeplabv2} & 25.89 & +0.00 \\ \midrule
		\multirow{5}*{\makecell[l]{Style and Fog\\ Decomposition}} & $\model_{s \rightarrow \imd}$ & $\model_{\imd \rightarrow \td}$ & $\model_{s \rightarrow \td}$ & mIoU & gain \\ \cline{2-6}
		&\cmark &          &         & 39.16 & +13.27  \\
		&\cmark & \cmark   &         & 42.49 & +16.60 \\
		&       &         & \cmark   & 40.21 & +14.32 \\
		&\cmark & \cmark  &  \cmark  & 43.06 & +17.17 \\ \midrule
		\multirow{4}*{Cyclical Training} & $T=1$ & $T=2$ & $T=3$ & mIoU & gain\\ \cline{2-6}
		&\cmark   &          &         & 45.32 & +19.43  \\
		&         & \cmark   &         & 45.78 & +19.89 \\
		&         &          & \cmark   & 45.45 & +19.56 \\ \midrule
		\multirow{4}*{Cumulative Loss} & $L1$ & $Cosine$ & $L2$ & mIoU & gain\\ \cline{2-6}
		&\cmark   &          &          & 47.64 & +21.75 \\
		&         & \cmark   &          & 47.23 & +21.34 \\
		&         &          & \cmark   & 48.21 & +22.32 \\ 
        \bottomrule
		\end{tabular}
	}
	\vspace{-4mm}
\end{table}

\subsection{Performance Comparison}

We compare our method against several kinds of methods, including
% backbone models
\textbf{1)}~\underline{\emph{backbones}}: RefineNet~\cite{lin2017refinenet} and DeepLab-v2~\cite{chen2017deeplab}; 
\textbf{2)}~\underline{\emph{defogging-based}}: MSCNN~\cite{ren2016single}, DCP~\cite{he2010single}, Non-local~\cite{berman2016non}, DCPDN~\cite{zhang2018densely} and GFN~\cite{ren2018gated};
% , and DCPDN~\cite{zhang2018densely};
\textbf{3)}~\underline{\emph{DA-based}}: Multi-task~\cite{alshammari2020competitive}, AdSegNet~\cite{tsai2018learning}, ADVENT~\cite{vu2019advent}, CCM~\cite{li2020content}, SAC\cite{araslanov2021self}, ProDA\cite{zhang2021prototypical}, DMLC~\cite{guo2021metacorrection}, DACS\cite{tranheden2021dacs} and our baseline DISE~\cite{chang2019all}; 
\textbf{4)}~\underline{\emph{synthesis-based}}: SFSU\cite{sakaridis2018semantic}, CMAda2\cite{sakaridis2018model}, CMAda3+\cite{dai2020curriculum}, CycleGAN\cite{zhu2017unpaired}, MUNIT\cite{huang2018multimodal}, AnalogicalGAN\cite{gong2021analogical}. 
\textbf{5)}~\underline{\emph{Defogging/Synthesis + DA-based}}: MSCNN+DISE, SFSU+DISE.
The mean Intersection-Over-Union (mIoU) results on Foggy Zurich and Foggy Driving are reported in Table~\ref{tab:sota_1}.

% 讲各类对比方法具体怎么操作
For \emph{Defogging-based} methods, we first use these methods to defog the real foggy test images and then use the backbone segmentation model to produce the predictions.
For \emph{Domain Adaptation based} methods, we set the source domain data of as clear cityscapes, $s$ domain of our method. As for the target domain data, we combine the Clear Zurich and Foggy Zurich, which are used as the $m$ domain and $t$ domain in our method. 
By using the same amount of training data, we ensure a fair comparison with \emph{DA-based} methods.
For \emph{Defogging}+\emph{Domain Adaptation} methods, we first use the MSCNN~\cite{ren2016single} to defog the target domain data (including training data and test data) and then use DISE~\cite{chang2019all} to bridge the domain gap.

For \emph{Synthesis-based} methods, the paradigm is to finetune the segmentation model pretrained on the real clear weather images (Cityscapes) with synthetic foggy images, e.g., Foggy Cityscapes DBF, and labels corresponding to its clear weather images. The difference in these \emph{Synthesis-based} methods is that they use different methods~\cite{sakaridis2018semantic,sakaridis2018model,gong2021analogical,zhu2017unpaired,huang2018multimodal} to generate the synthetic foggy images. Finally, the finetuned model is tested on real foggy images.
For a fair comparison with CMAda3+~\cite{dai2020curriculum}, we also add Foggy Cityscapes DBF as extra data to train sub-network $\model_{m \rightarrow \td}$ before cumulative training, which we name as CuDA-Net+.

% 
% The results in Table~\ref{tab:sota_1} show that although the backbone model DeepLab-v2 performs not well as RefineNet, our proposed method CuDA-Net (using DeepLab-v2 as the backbone) achieves a top performance, outperforming all state-of-the-art methods. 
% Furthermore, we can see that the domain adaptation methods, which adapt segmentation model from domain $s$ directly to domain $\td$, can not achieve significant improvement compared to our method. This is consistent with our assumption that general domain adaptation method can not perform well when the domain gap are too large and affected by different factors (style and fog), which also prove the necessity of investigating both style factor and fog factor in this setting.
% The results also show that the defogging based methods can not always obtain good performances. It is because defogging based methods require pair-wised training samples to remove the fog, but we can not obtain this kind of training data in SFSU.

The results in Table~\ref{tab:sota_1} show that although the backbone model DeepLab-v2 performs not well as RefineNet, our proposed method CuDA-Net (using DeepLab-v2 as the backbone) achieves a top performance, outperforming all state-of-the-art methods. 
We also achieve SOTA on ACDC~\cite{ACDC} dataset (see \href{https://acdc.vision.ee.ethz.ch/benchmarks#semanticSegmentation}{ACDC-fog benchmark website}).
% \emph{supplementary material}
Besides, we can see that the DA-based methods, which directly adapt the segmentation model from domain $s$ to domain $\td$, can not significantly improve the performance compared to our method. This is consistent with our assumption that general domain adaptation methods can not perform well when the domain gap is too large and affected by different factors (style and fog), also proving the necessity of investigating both style and fog factor in this setting.
The results also show that the defogging-based methods cannot always obtain good performances. It is because defogging-based methods require pair-wise training data to remove the fog, but we can not obtain this kind of data in SFSU.

% Also, this may be due to the mismatch between the objectives of defogging and segmentation.
%Note that, in Table~\ref{tab:images_to_use}, our proposed method ADA
%just a little weak compared with the state-of-the-art CMAda3+(46.8\% on \textsc{Foggy Zurich}). 
% \noindent\textbf{Training data comparison with CMAda3+.}
In Table~\ref{tab:sota_1}, when we introduce the synthetic foggy scene datasets --- Foggy Cityscapes DBF simulated in CMAda3+ to our method, our CuDA-Net+ further improve the performance, outperforming CMAda3+ by 2.3\% on FZ (3.7\% on FD). 
Note that, our CuDA-Net+ improves by 0.9\% from CuDA-Net on FZ when we only introduce 498 dense synthetic foggy images, indicating synthesizing foggy images and our CuDA-Net can complement each other very well. 
However, combining DISE~\cite{chang2019all} with the defogging method MSCNN~\cite{ren2016single} or the fog synthesis method SFSU~\cite{sakaridis2018semantic} cannot yield better performance than only using DISE~\cite{chang2019all}.

The qualitative comparison is shown in Figure~\ref{fig:sota}. 
The red boxes clearly show that our method CuDA-Net can better deal with the details than CMAda3+, especially for the classes in the boundary of sky and other objects.

% We also combine the DISE with the defogging method MSCNN or the fog synthesis method SFSU and these two methods are not as good as only using DISE, which shows defogging and fog synthesis will introduce other domain gap in this problem.

%\textcolor{red}{Although our method does not use depth information to generate foggy images independently and also exploits a relatively weak baseline DeepLab-v2, our method achieves considerable performances, compared with CMAda2. It shows that the proposed method might be capable of simulating the fog synthesis process.}

\begin{figure*}[!ht]
\centering
		\tabcolsep=0.5pt
		\renewcommand\arraystretch{0.5}
		\begin{tabular}{cccccccc}
			\includegraphics[width=0.14\textwidth]{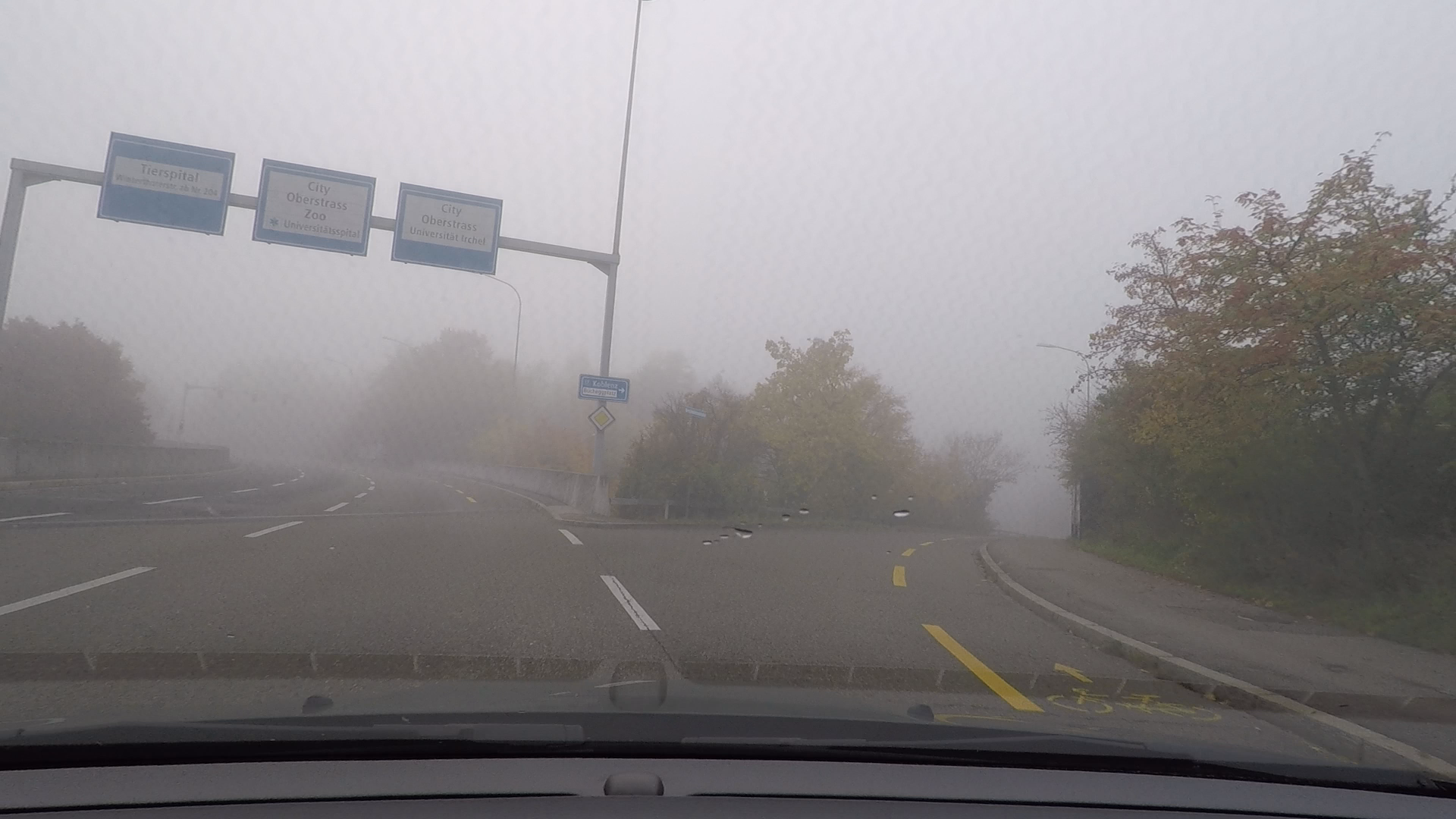} &
			\includegraphics[width=0.14\textwidth]{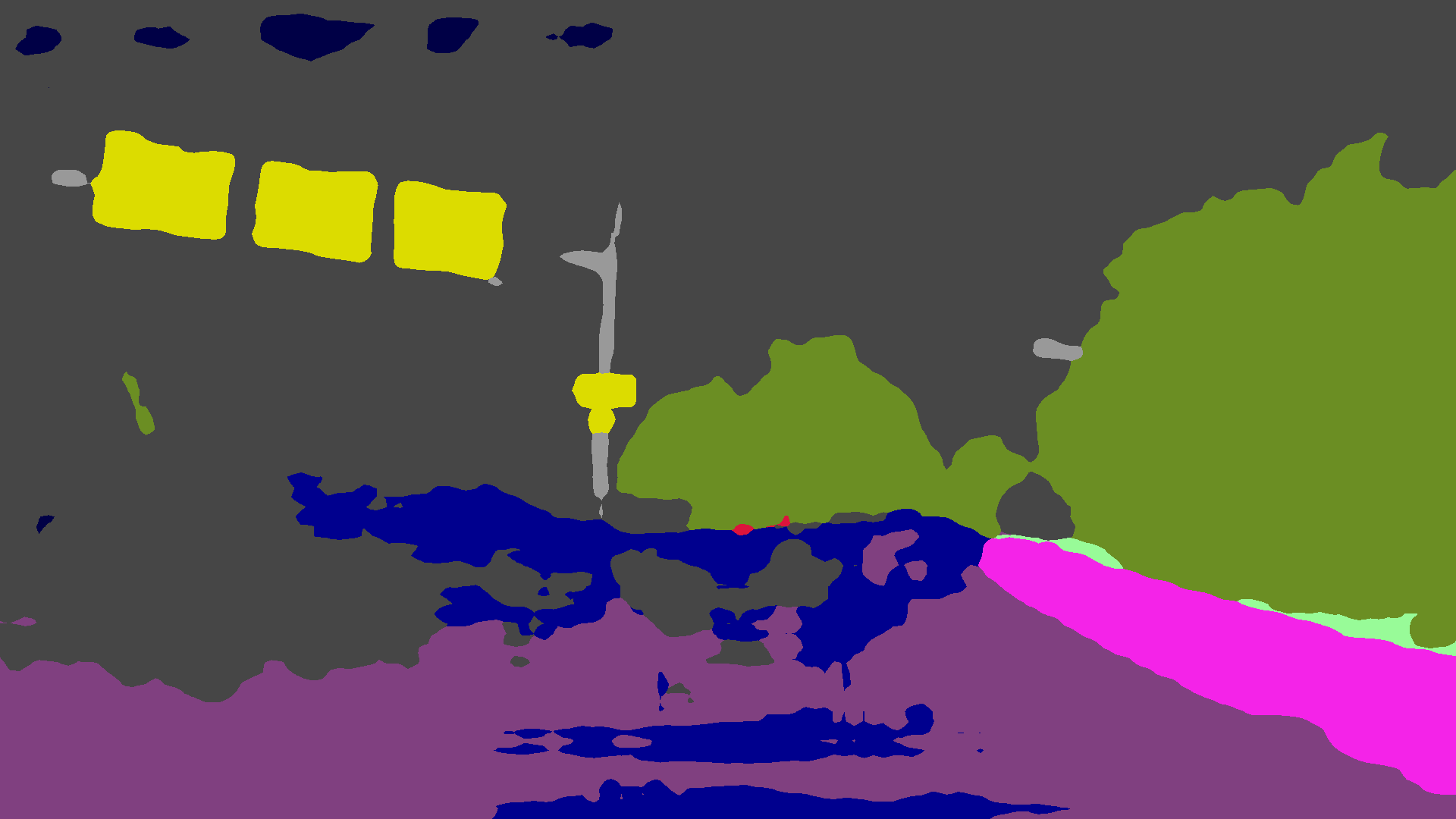} &
			\includegraphics[width=0.14\textwidth]{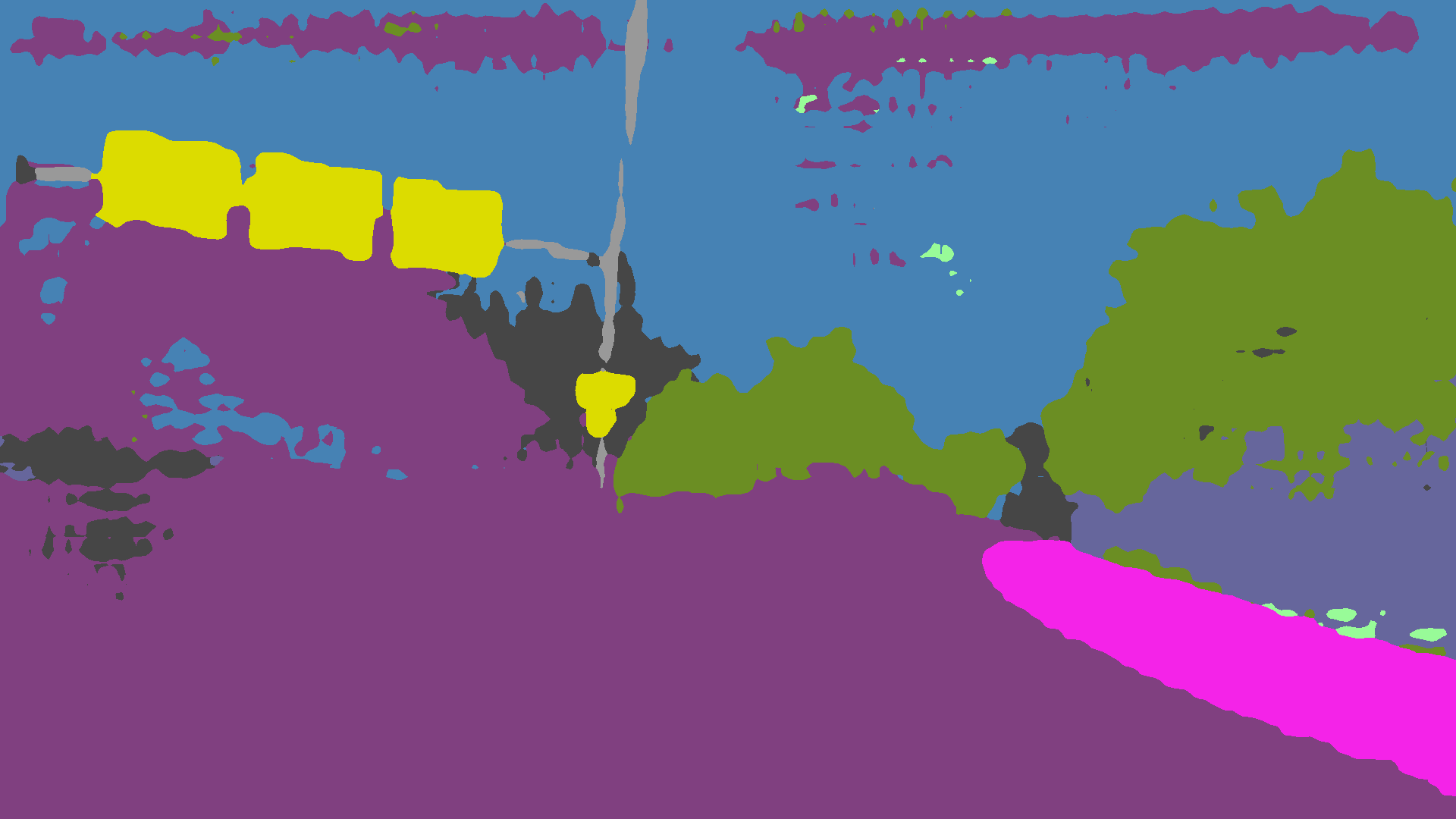} &
			\includegraphics[width=0.14\textwidth]{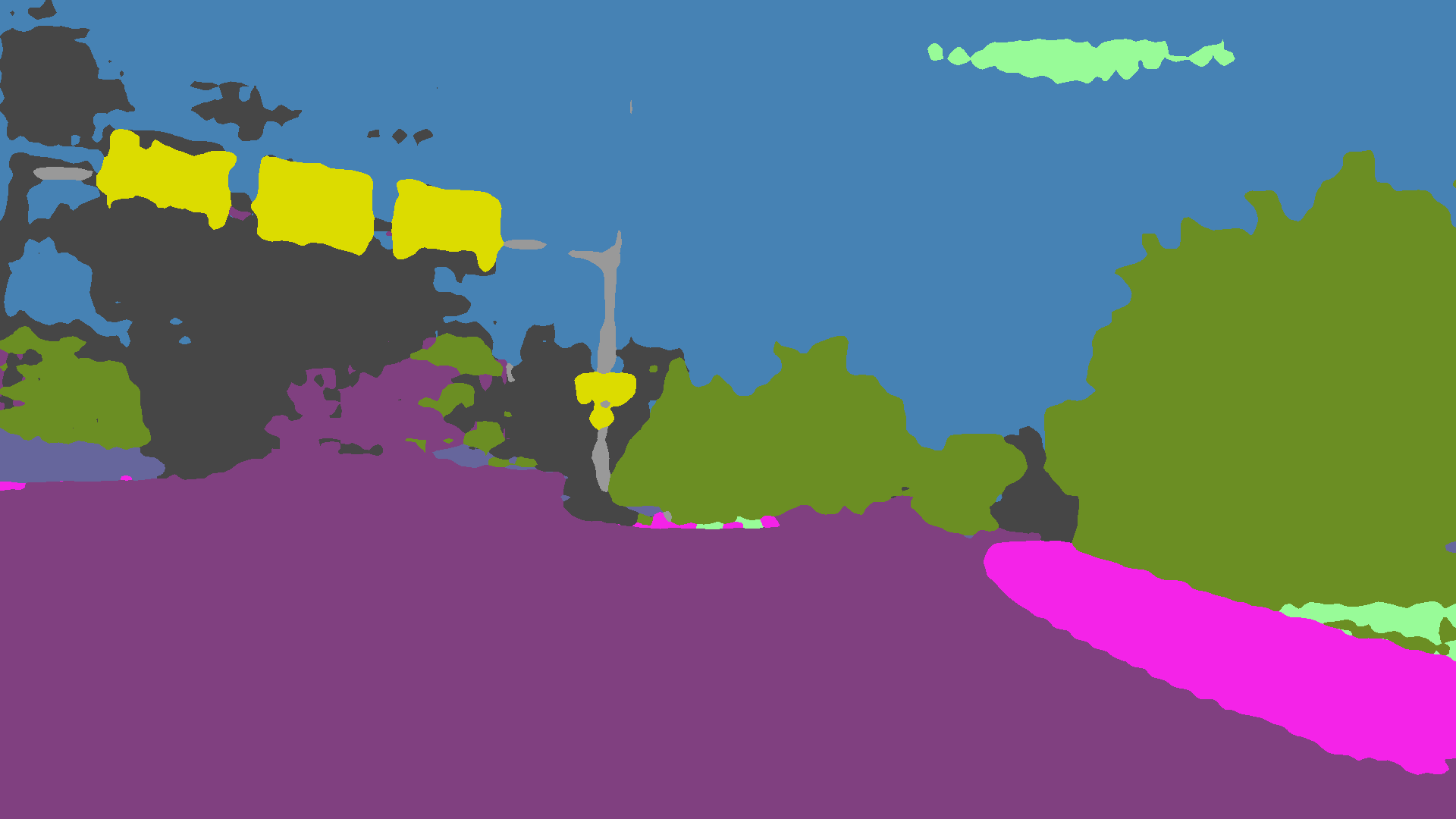} &
			\includegraphics[width=0.14\textwidth]{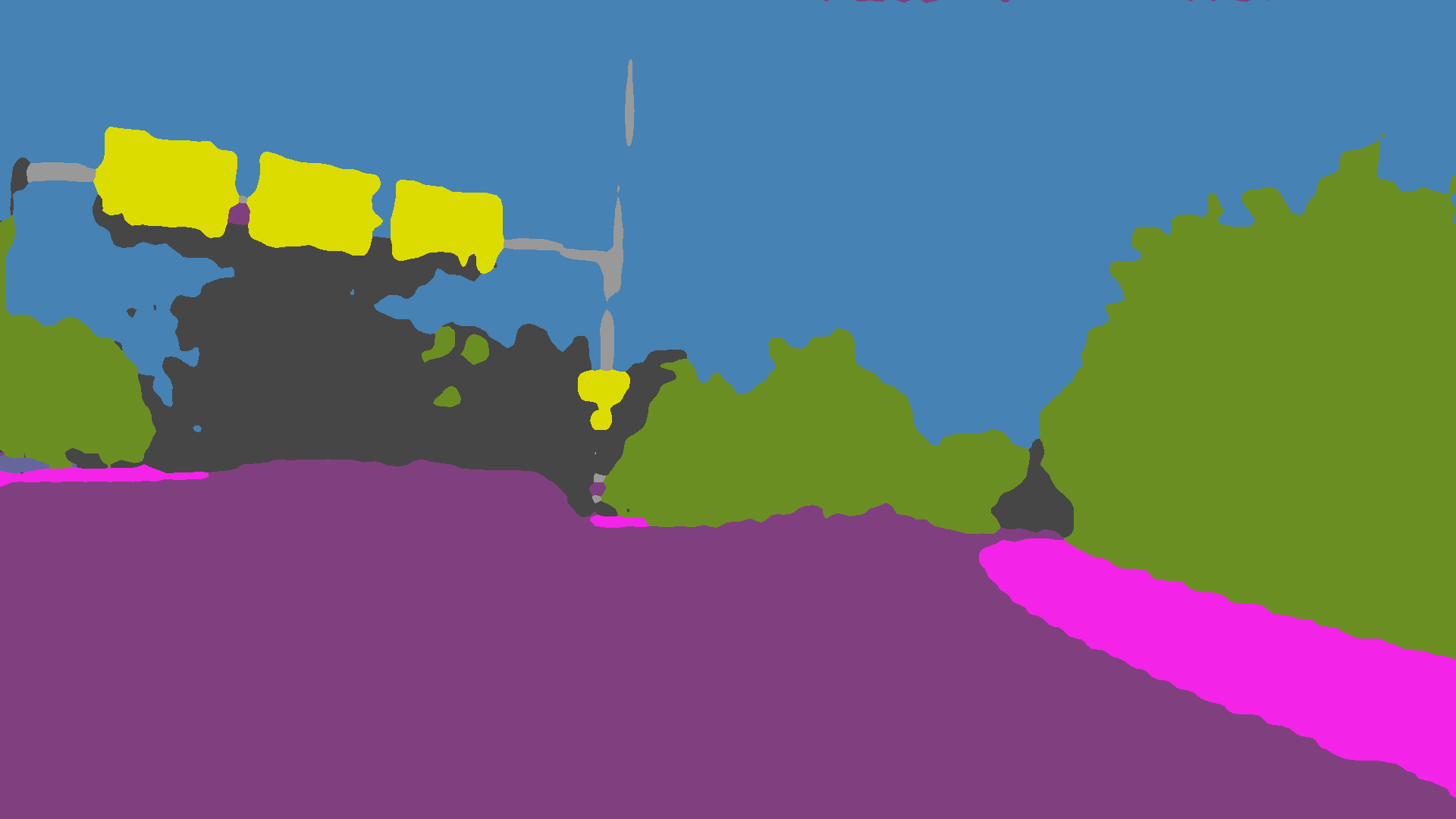} &
			\includegraphics[width=0.14\textwidth]{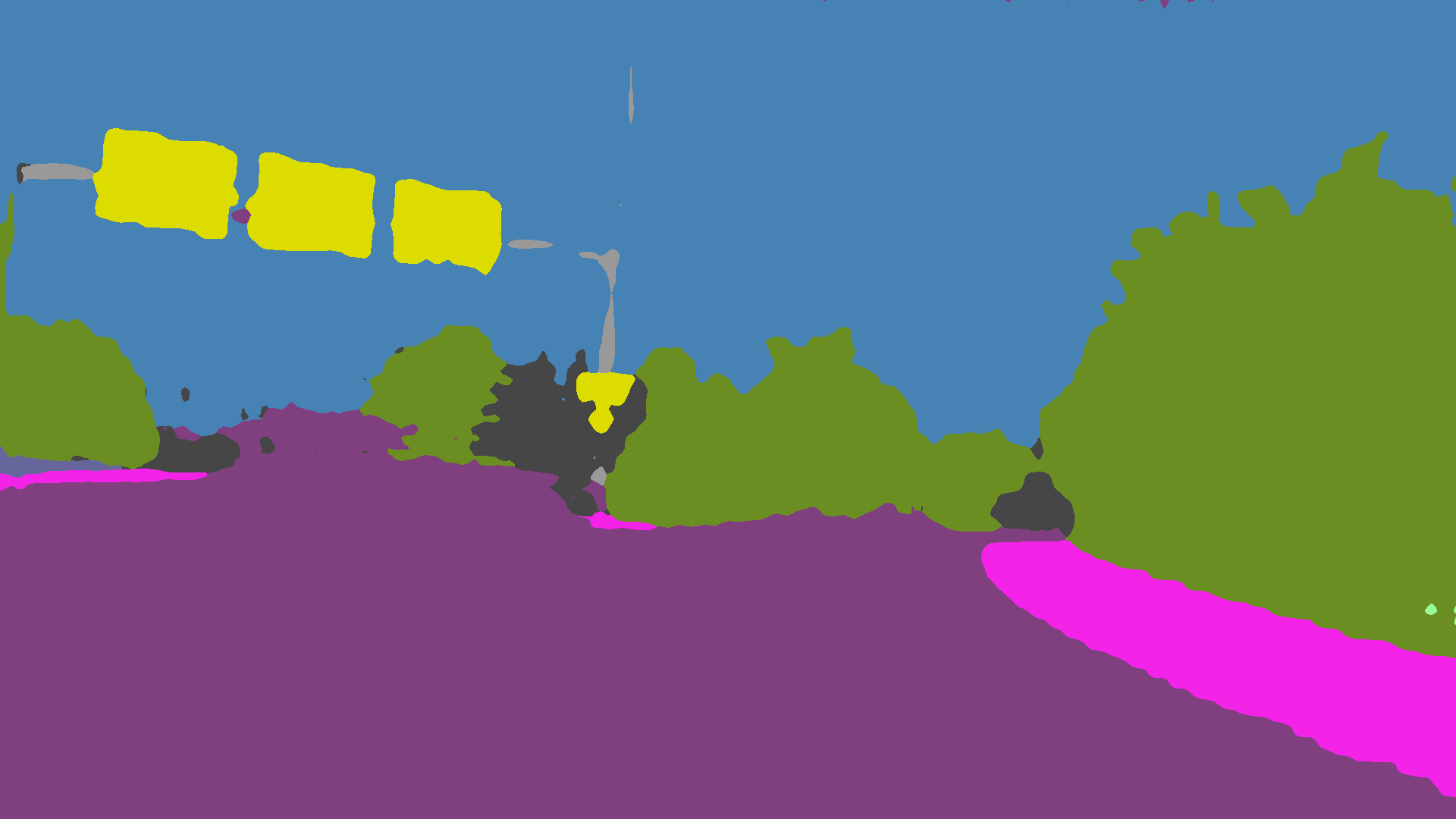} &
			\includegraphics[width=0.14\textwidth]{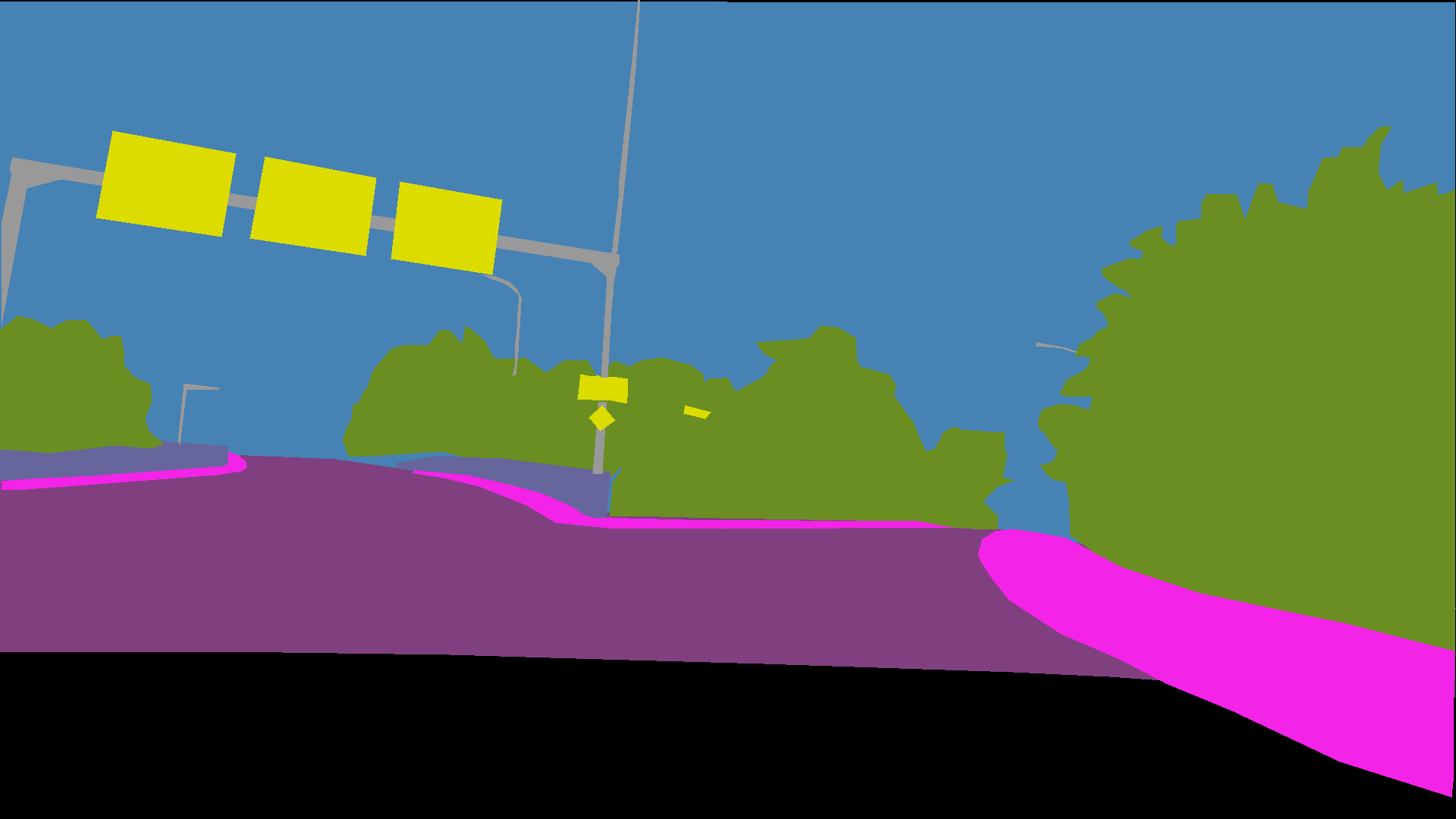}\\
			\includegraphics[width=0.14\textwidth]{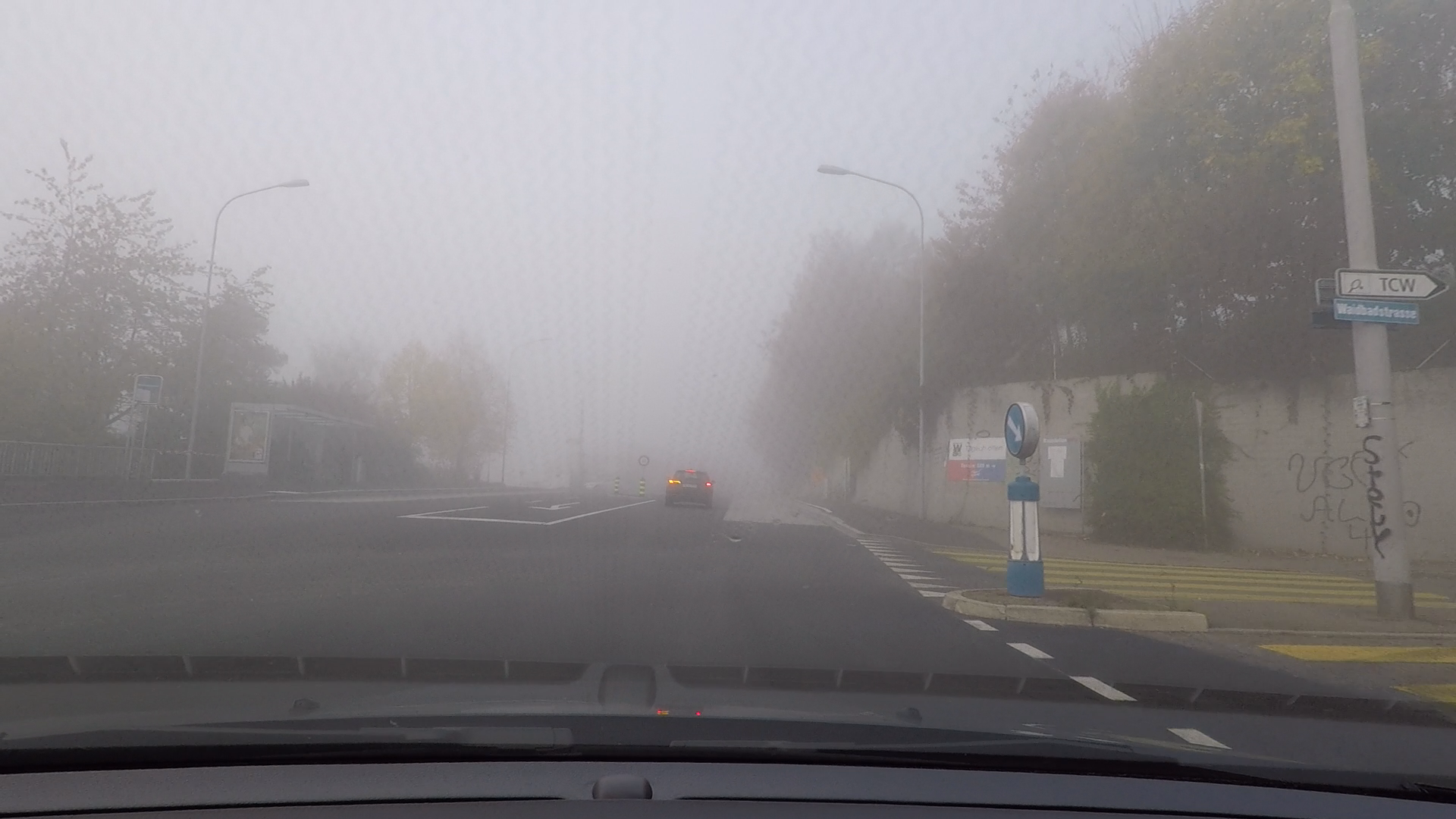} &
			\includegraphics[width=0.14\textwidth]{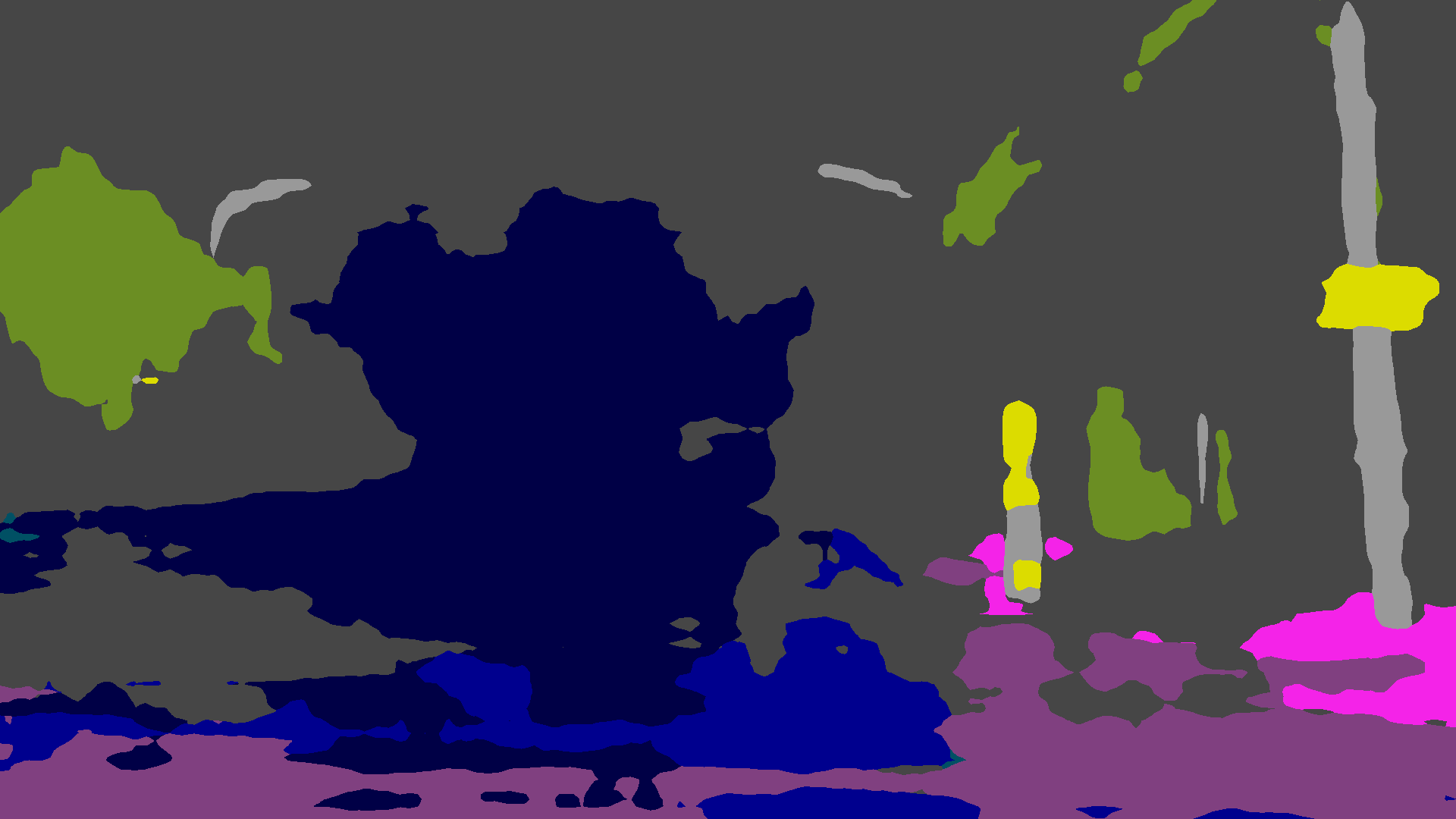} &
			\includegraphics[width=0.14\textwidth]{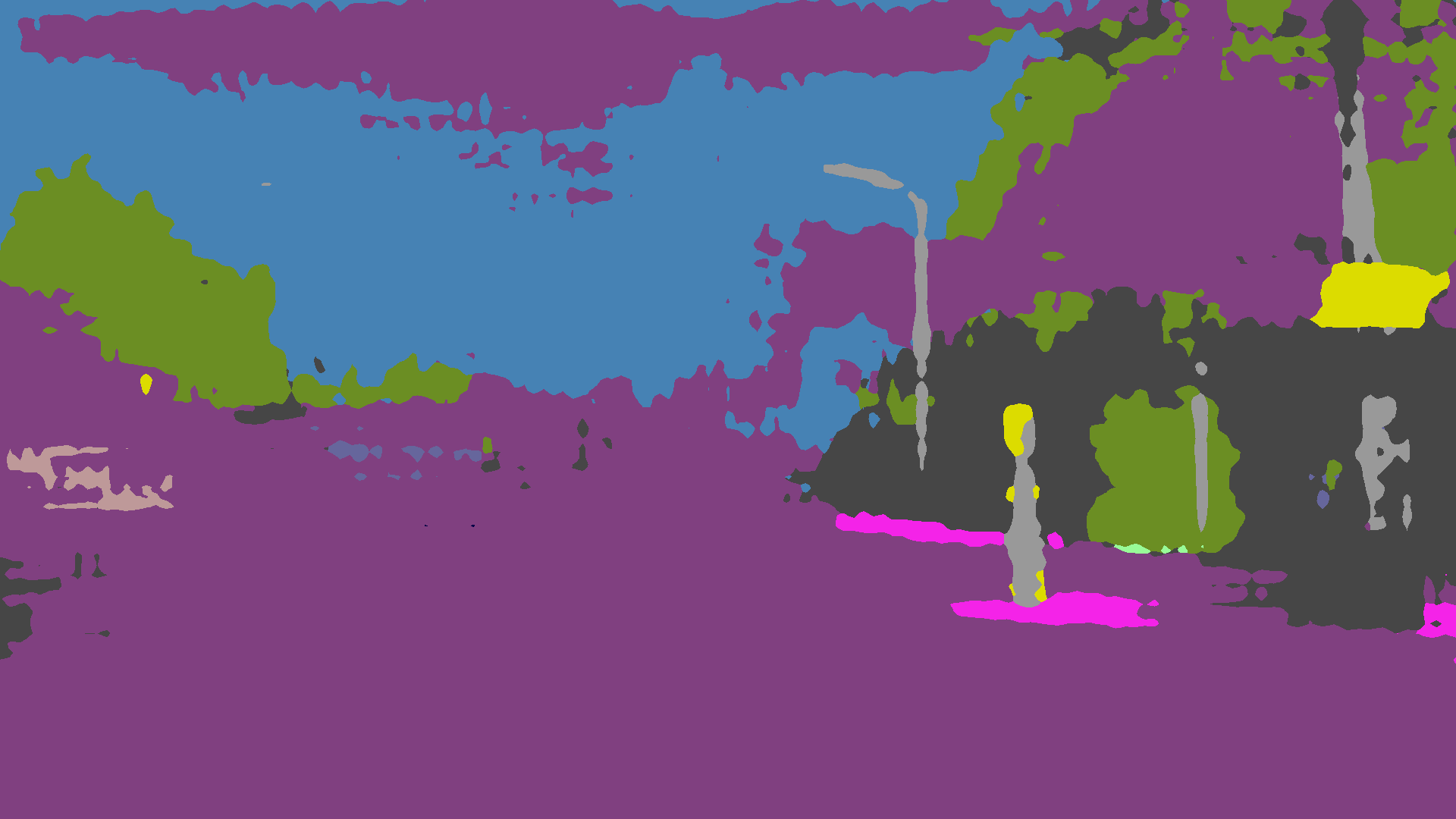} &
			\includegraphics[width=0.14\textwidth]{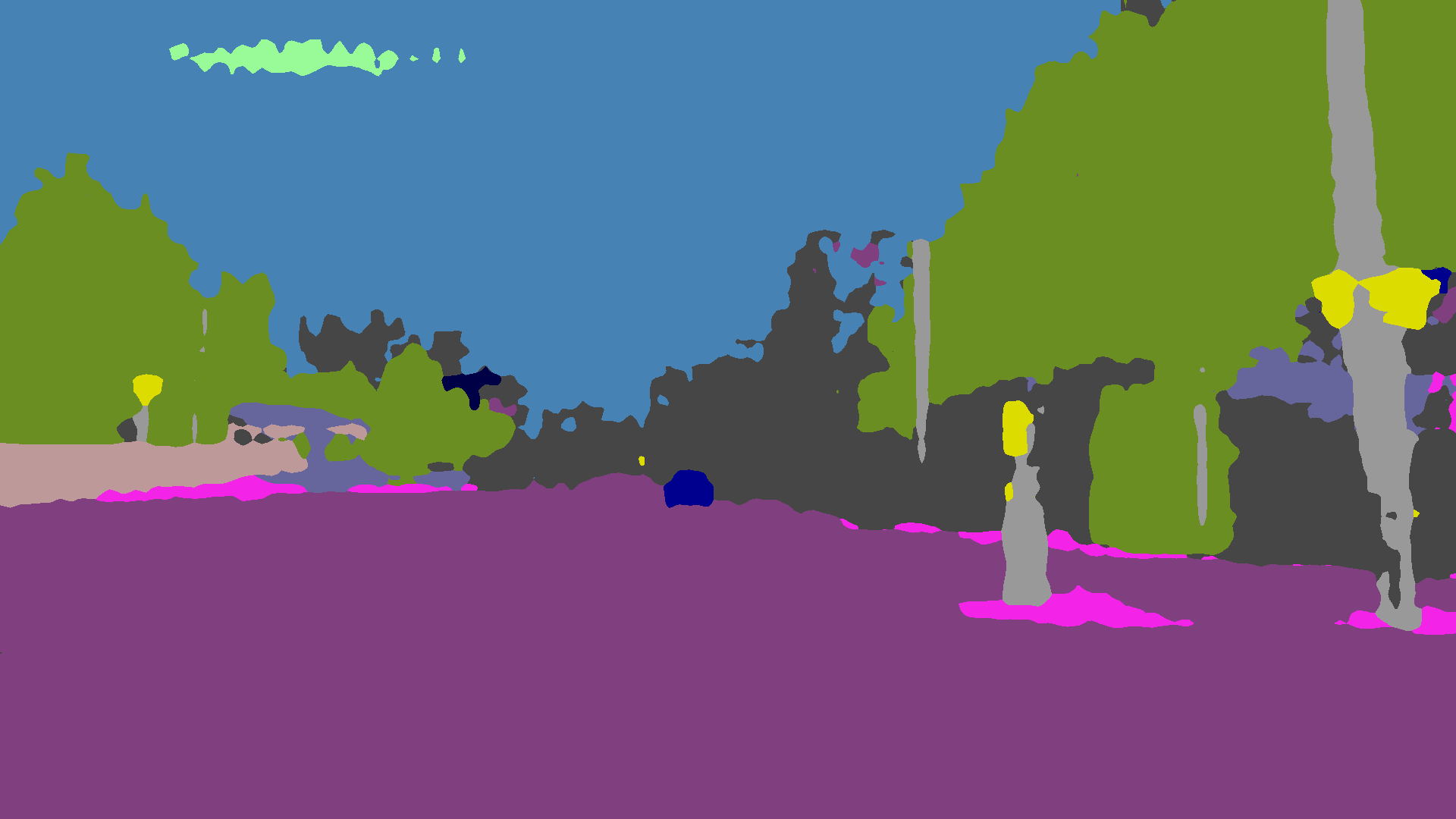} &
			\includegraphics[width=0.14\textwidth]{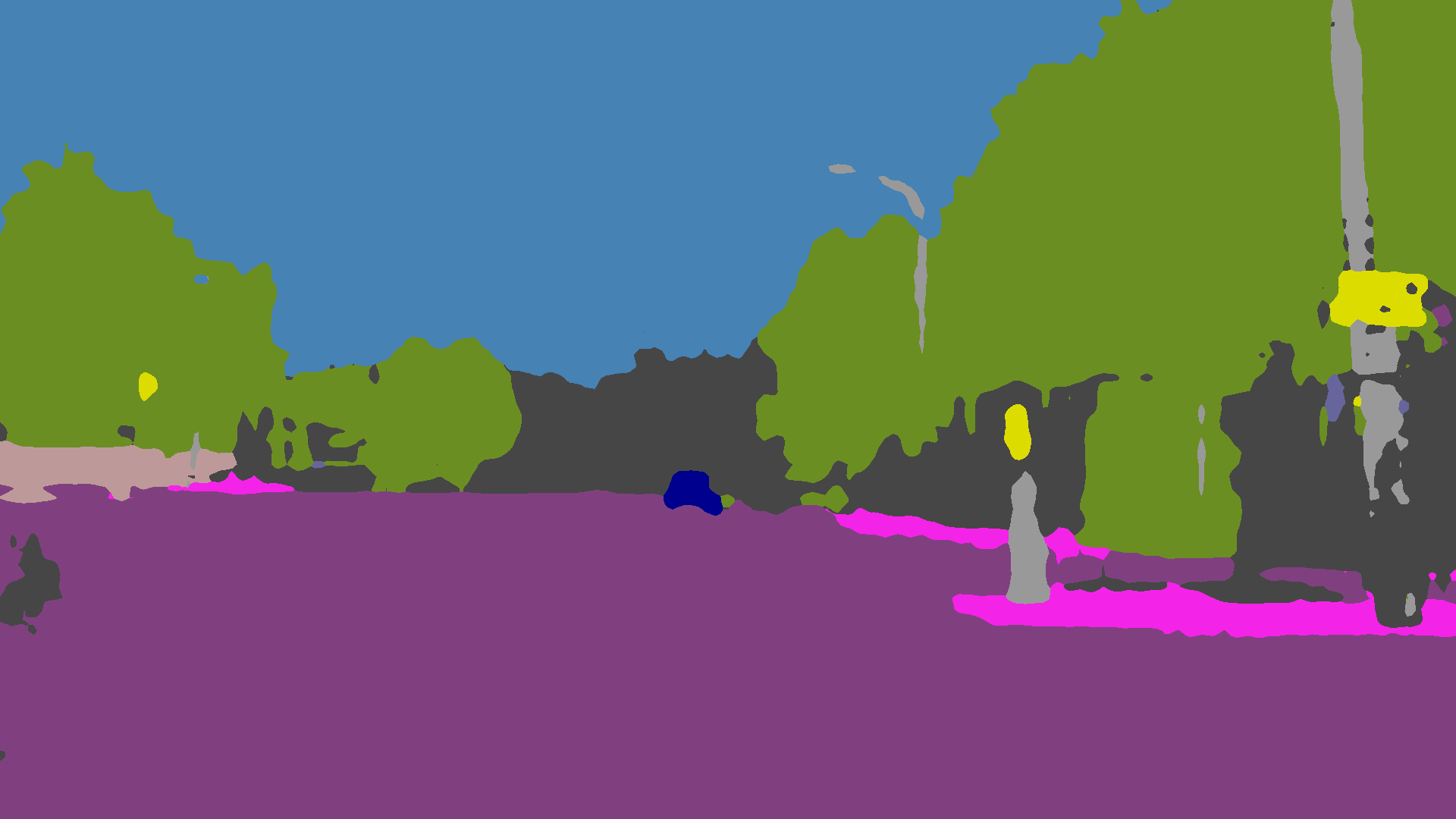} &
			\includegraphics[width=0.14\textwidth]{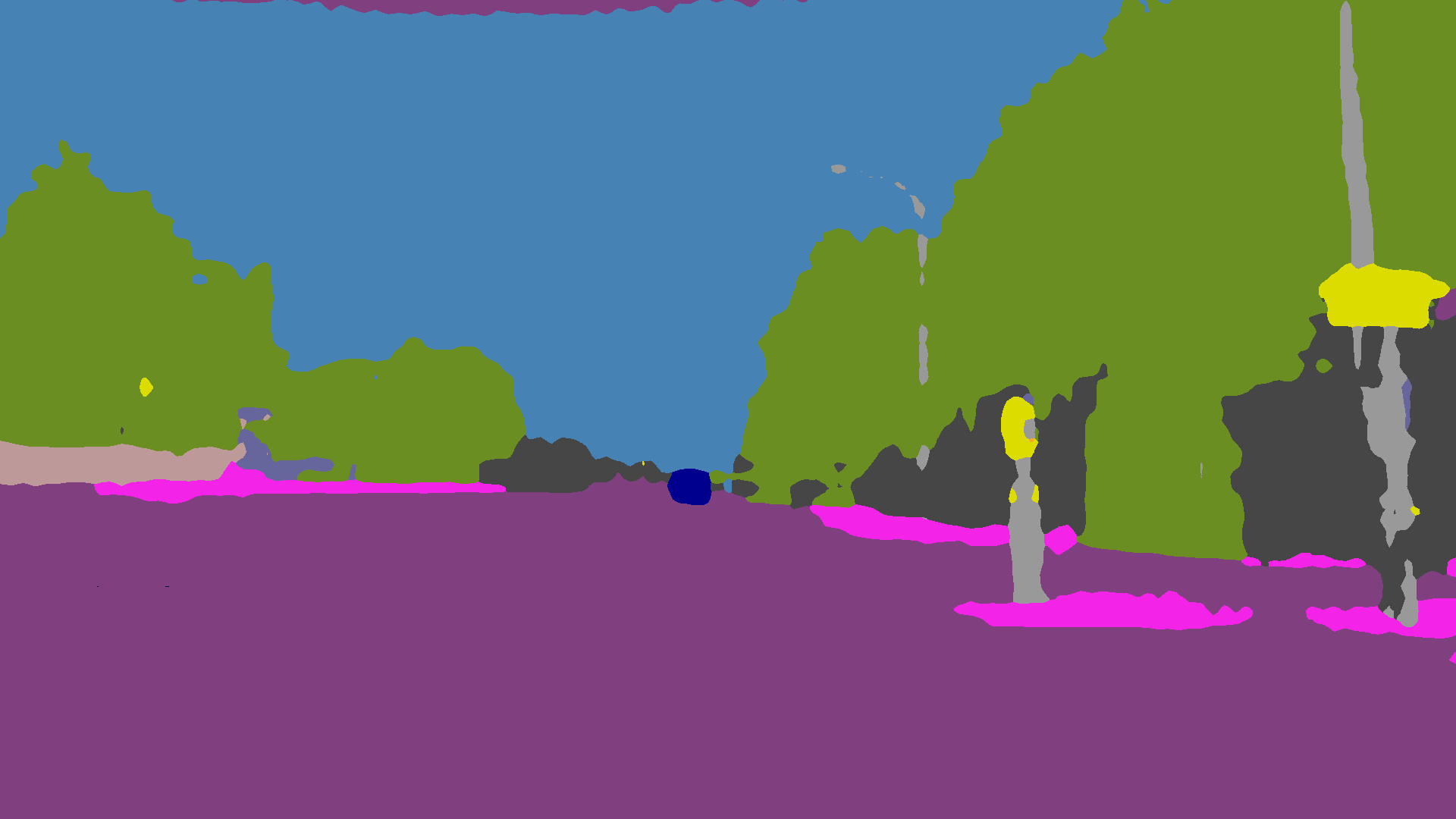} &
			\includegraphics[width=0.14\textwidth]{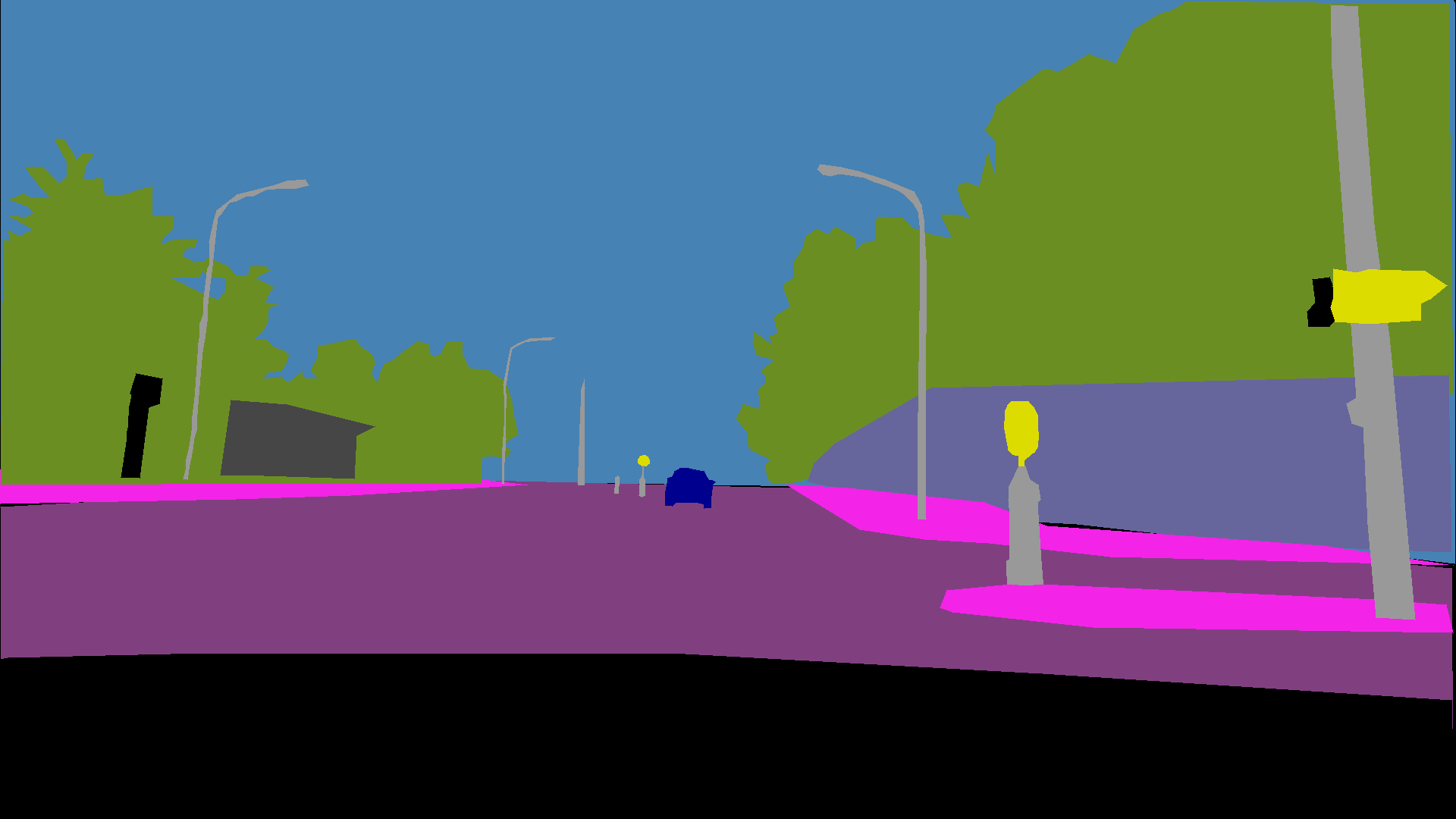}\\
			\includegraphics[width=0.14\textwidth]{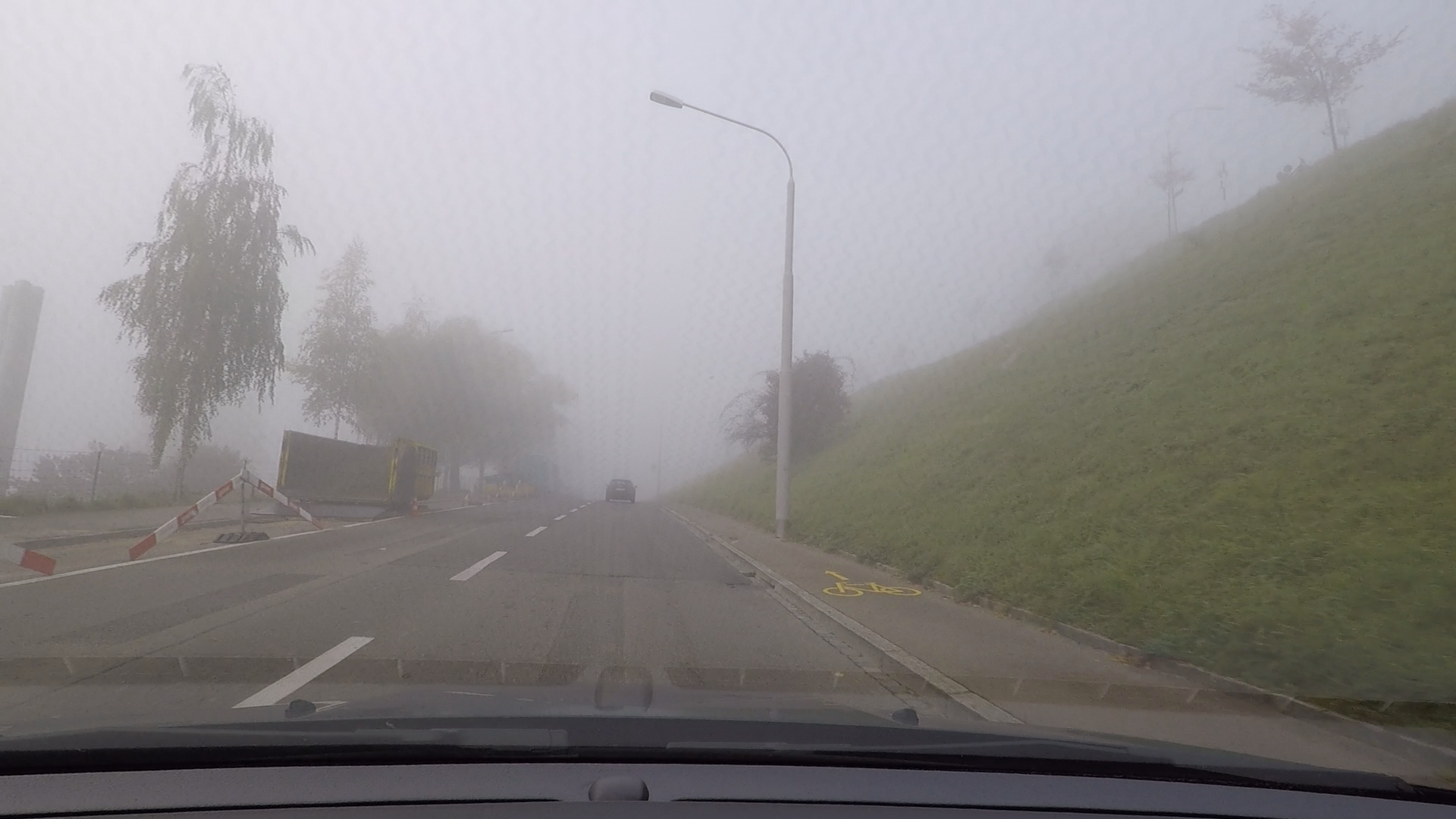} &
			\includegraphics[width=0.14\textwidth]{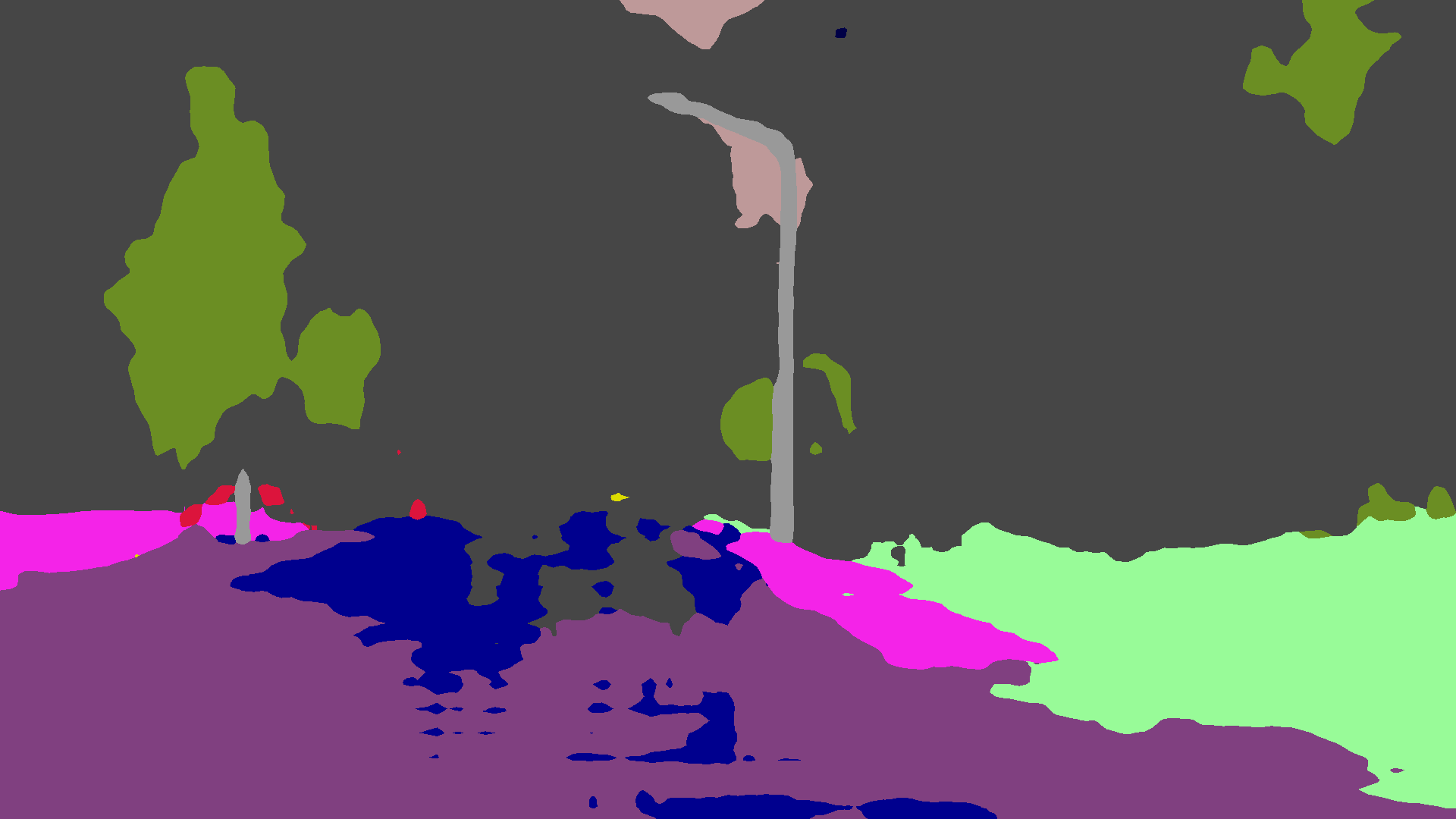} &
			\includegraphics[width=0.14\textwidth]{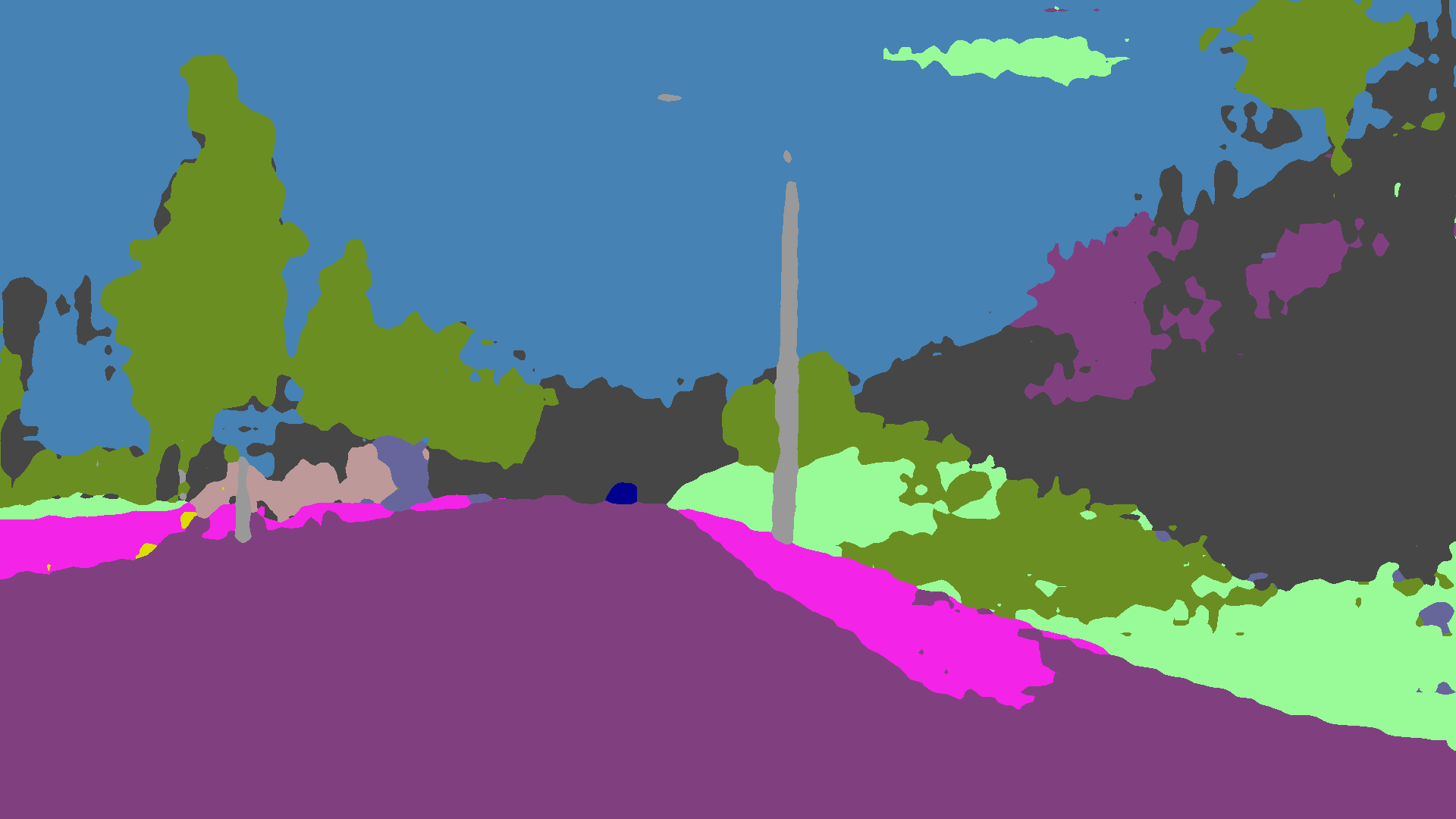} &
			\includegraphics[width=0.14\textwidth]{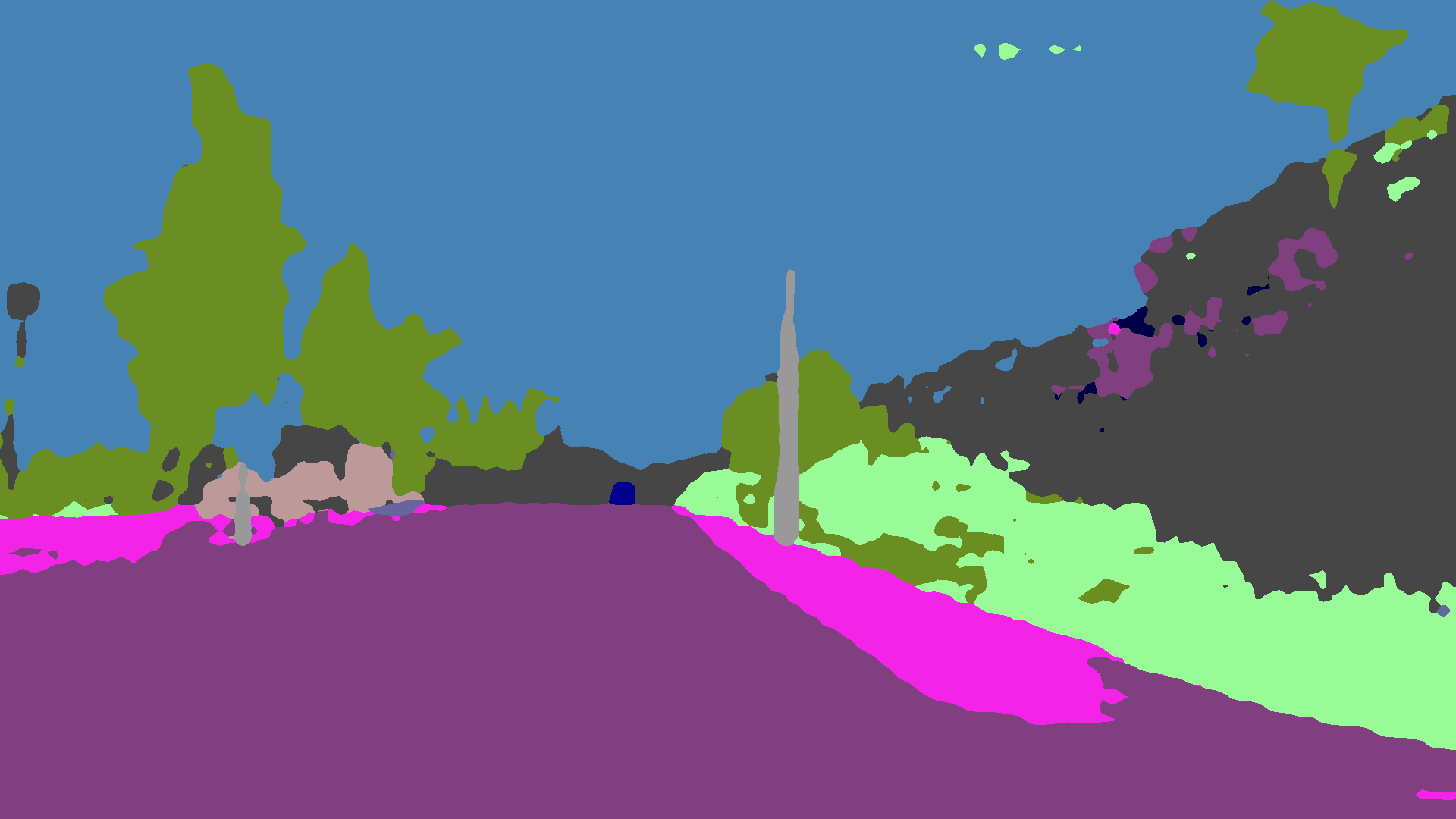} &
			\includegraphics[width=0.14\textwidth]{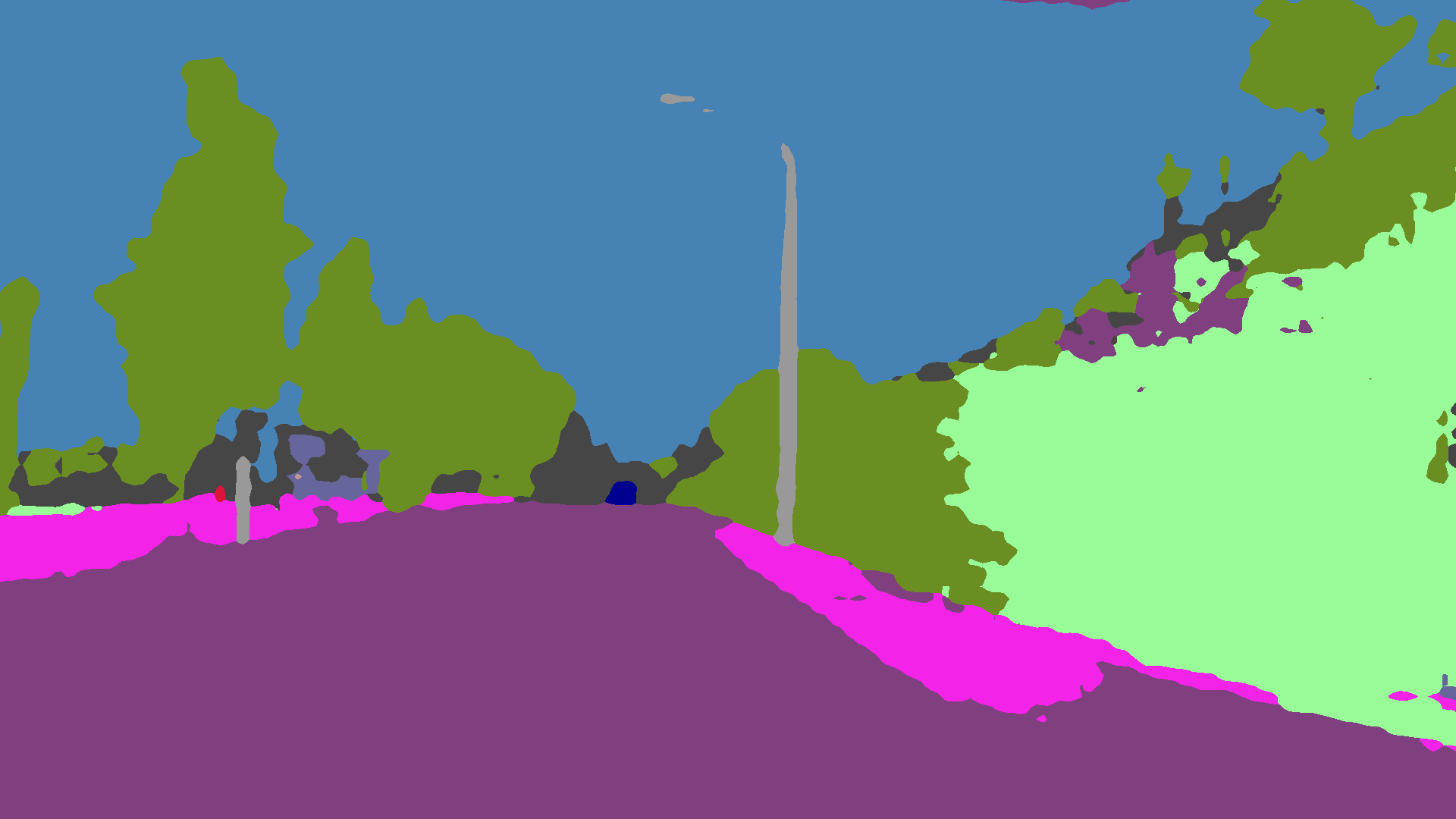} &
			\includegraphics[width=0.14\textwidth]{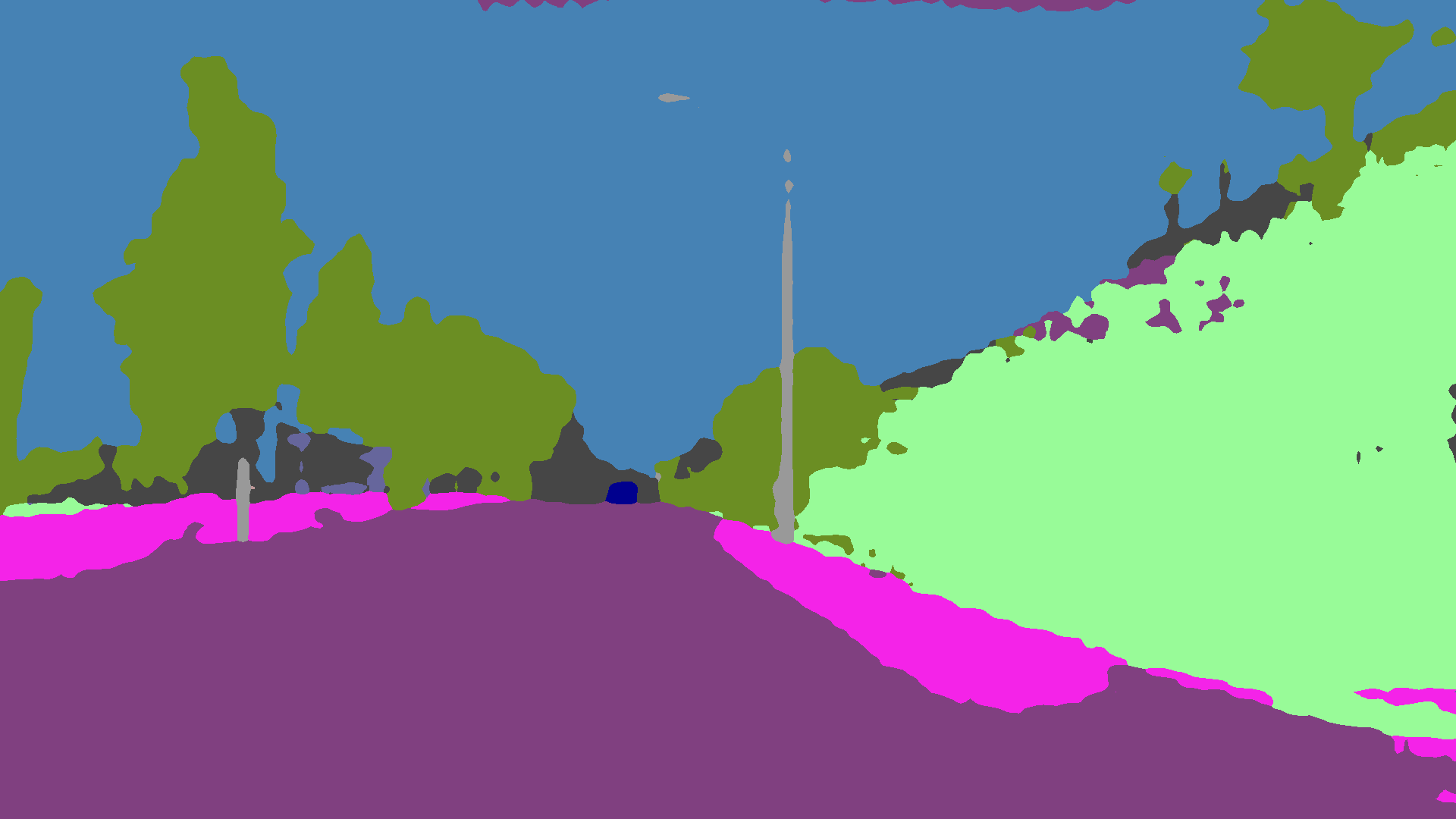} &
			\includegraphics[width=0.14\textwidth]{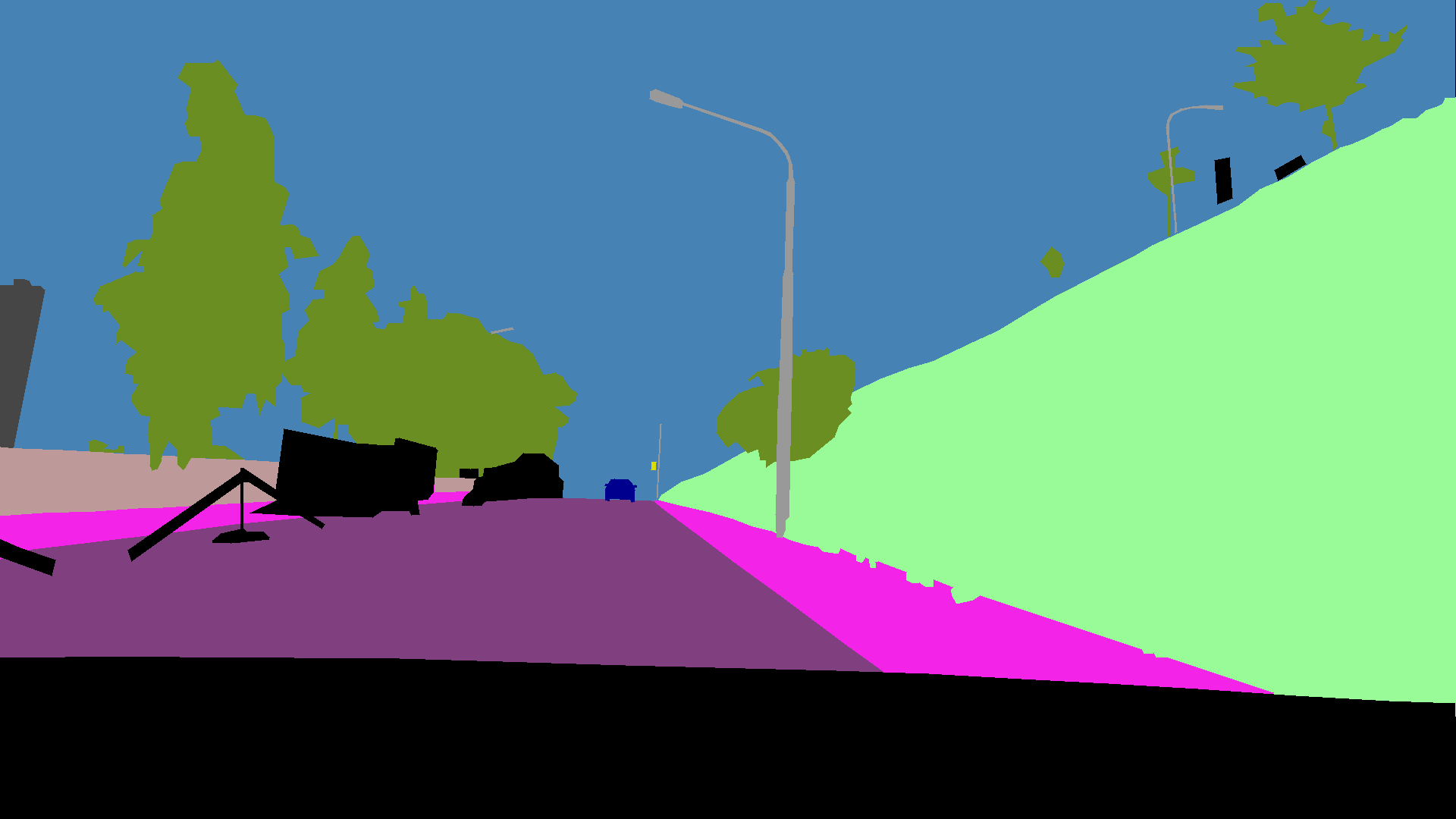}\\
			Input & DeepLab-v2 & 
			\small{+$\model_{s \rightarrow \imd}$} & \small{+$\model_{s \rightarrow \imd}$+$\model_{\imd \rightarrow \td}$} & \small{+$\model_{s \rightarrow \imd}$+$\model_{\imd \rightarrow \td}$} & CuDA-Net  & Ground Truth \\
			\scriptsize{} & \scriptsize{} & 
			\scriptsize{} & \scriptsize{} & \small{+$\model_{s \rightarrow \td}$} 
			& \scriptsize{} & \scriptsize{} \\
	\end{tabular}
	\vspace{-2mm}
	\caption{
% 	Some subjective segmentation results of the proposed methods
    \textbf{Qualitative results of ablation study}.
	These experiments are conducted on the Foggy Zurich-test dataset. Each column shows the results of the proposed method with different components. 
	The results show more clear spatial structure as more components are used.}
	\vspace{-4mm}
	\label{fig:ablation}
\end{figure*}

\begin{table}[!t]
\begin{center}
\caption{\textbf{Different selection schemes for constructing $\imd$ domain.} We set three random seeds to ensure fair comparison and test the trained model on Foggy Zurich-test dataset.}
\label{tab:selection}
\vspace{-2mm}
\resizebox{0.8\linewidth}{!}{
\begin{tabular}{l|ccccc}
\hline
Random seed             & 10     &   & 20    &    & 50      \\ \hline
CNN-based selection     & 47.5   &  & 47.5   &    & 47.2       \\ \hline
Manual selection        & 48.0   &  & 48.2   &    & 48.2        \\ \hline
\end{tabular}}
\vspace{-6mm}
\end{center}
\end{table}

\begin{figure}[!ht]
\centering
		\tabcolsep=0.5pt
		\renewcommand\arraystretch{0.5}
		\begin{tabular}{cccccccc}
			\includegraphics[width=0.16\textwidth]{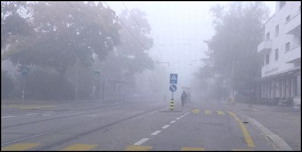} &
			\includegraphics[width=0.16\textwidth]{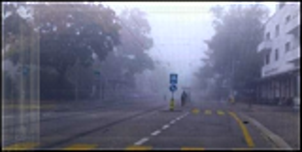} &
			\includegraphics[width=0.16\textwidth]{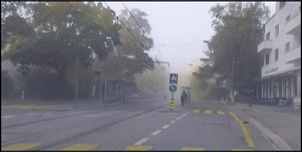} 
			
			\\
			\includegraphics[width=0.16\textwidth]{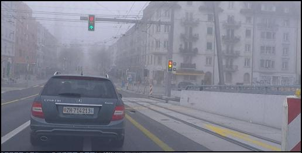} &
			\includegraphics[width=0.16\textwidth]{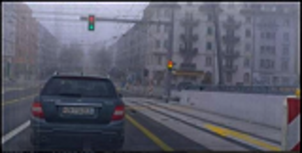} &
			\includegraphics[width=0.16\textwidth]{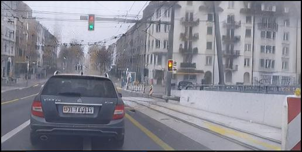} 
			 \\
			Input & GFN &  Ours ($\model_{\imd \rightarrow \td}$) \\
	\end{tabular}
	\vspace{-2mm}
    \caption{\textbf{The ability of defogging}. We compare our defogged images generated by the $\model_{\imd \rightarrow \td}$ in CuDA-Net with those from the conventional defogging method GFN~\cite{ren2018gated}. The input images are randomly selected from {Foggy Zurich}.}
    \vspace{-4mm}
    \label{fig:defog}
\end{figure}

\subsection{Discussion}
In this section, we conduct a series of ablation studies to validate the contributions of individual components to the final foggy scene understanding. 
% As mentioned previously, our proposed CuDA-Net includes three sub-networks and a cumulation relationship constraint to make three networks fine-tune collaboratively. Therefore, 

\noindent\textbf{Effectiveness of style and fog decomposition.} In Table~\ref{tab:ablation}, the non-adapted model Deeplabv2, which is also the backbone of our CuDA-Net, only gives 25.89 mIoU on FZ. When using `$\model_{s \rightarrow \imd}$', the performance increases to 39.16, revealing that the style adaptation matters. When using `$\model_{s \rightarrow \imd}$+$\model_{\imd \rightarrow \td}$', i.e. first conducting style adaptation and then fog adaptation, we bring +3.33 mIoU gain, additionally indicating that the fog adaptation matters.
Note that using `$\model_{s \rightarrow \imd}$+$\model_{\imd \rightarrow \td}$' is 2.28 higher than only using `$\model_{s \rightarrow \td}$', which demonstrates that directly transferring with style and fog is not as good as two step adaptation. Other than that, using `$\model_{s \rightarrow \imd}$+$\model_{\imd \rightarrow \td}$+$\model_{s \rightarrow \td}$' boosts the performance to 43.06, showing the necessity of dual-factor adaptation.

\noindent\textbf{Effectiveness of cyclical training.} We investigate the importance of cyclical training without ${L}_{cum}$, \ie using Equation (\ref{equa:stage1_loss}). As shown in Table~\ref{tab:ablation}, when we set $T$ as 2, cyclical training improves the performance by 2.72. When we set $T$ as 1 or 3, both results are close to 45.78, which shows the performance is not sensitive to the selection of $T$.

\noindent\textbf{Effectiveness of cumulative loss.} We also investigate the effects of cumulative loss ${L}_{cum}$ in Table~\ref{tab:ablation}. We fix the $T$ as 2 and use different distance metrics to calculate the domain discrepancy between two domains in the cumulative training. We find the L2 distance attains top performance.

We also show some subjective segmentation results in Figure~\ref{fig:ablation}. They clearly indicate that segmentation results go better as more components are used in CuDA-Net.

\noindent\textbf{Effects of the hyper-parameter $\lambda_{cum}$ in Equation (\ref{equa:stage2_loss}).} We conduct ablation study on $\lambda_{cum}$ on Foggy Zurich-test dataset in Figure~\ref{fig:lamda}. The results show our model is not sensitive to $\lambda_{cum}$ and we set $\lambda_{cum}$ as 0.25 because its top performance. We also conduct an loss ablation study on Equation (\ref{equa:stage1_loss}), results are shown in the \emph{supplementary material}.

\noindent\textbf{Manual selection v.s. CNN-based selection for constructing $\imd$ domain dataset.}
For constructing Clear Zurich, we manually select 248 images from the light category of Foggy Zurich*~\cite{sakaridis2018model} based on the human vision, to see whether they are clear or not. To prove its effectiveness, we also train a CNN to discriminate the clearness of the images in the light category of Foggy Zurich*~\cite{sakaridis2018model} and select the top 248 images to construct the $\imd$ domain. We set random seeds as 10, 20 and 50 to train the whole CuDA-Net framework. As shown in Table~\ref{tab:selection}, we find the manual selection functions better than CNN-based selection, which shows the necessity of our manual selection scheme.

% \subsection{Visualization of Defogging and Fog Synthesis}
% Although the proposed CuDA-Net aims to transfer style and fog for foggy scene understanding, it is also capable of conduct defogging on foggy images and synthesizing fog on clear images during disentanglement learning. In this subsection, we visualize the results of defogging and fog synthesis, and compare with conventional methods to show these abilities. 

\noindent\textbf{Visualization of Defogging.} 
Although our CuDA-Net aims to transfer style and fog for foggy scene understanding, it is also capable of defogging foggy images during disentanglement learning, as mentioned in the cross-domain translation part of Section~\ref{sec:FDN}. 
% In Figure~\ref{fig:defog}, we show the defogged images generated by the $\model_{\imd \rightarrow \td}$ in CuDA-Net and compare them with those from the conventional defogging method GFN~\cite{ren2018gated}.
In Figure~\ref{fig:defog}, we visualize the results of defogging and compare our method with the defogging method GFN~\cite{ren2018gated}.
% compare our visual results with the conventional defogging method GFN~\cite{ren2018gated}.
% to show this ability. 
The results clearly show that our method can remove the fog well and does not destroy the content of the images, while GFN~\cite{ren2018gated} brings in the color distortion. 
% We consider that it is because GFN requires pair-wise images (a clear one and a foggy one) to train the network, while our method does not have such kind of setting.

\begin{table}[t]
\begin{center}
\caption{\textbf{Generalization to rainy and snowy scenes.} We train our baseline on the ACDC rainy and snowy subsets and test it on the corresponding validation set, where $\model_{s \rightarrow \imd+\td}$ means we combine the $\imd$ domain and $\td$ domain data as the whole target domain data.}
\label{tab:generalization}
% \vspace{-2mm}
\resizebox{0.9\linewidth}{!}{
\begin{tabular}{l|ccc}
\hline
Setting    & $\model_{s \rightarrow \imd+\td}$   & $\model_{s \rightarrow \imd}$   & $\model_{s \rightarrow \imd}$+$\model_{\imd \rightarrow \td}$  \\ \hline
ACDC (rain)     & 46.2        & 43.9          & 48.5       \\ \hline
ACDC (snow)     & 44.8        & 42.6          & 47.2        \\ \hline
\end{tabular}}
\vspace{-6mm}
\end{center}
\end{table}

\noindent\textbf{Generalization to rainy and snowy scenes.} Thanks to the ACDC~\cite{ACDC} datasets, we can test our method on rainy and snowy scenes in Table~\ref{tab:generalization}. The results show that our proposed two-steps adaptation is better than directly adapting from the source domain to the target domain in other adverse scenes, indicating the potential of our method to address the understanding of different adverse scenes.

\section{Conclusion}
% For the SFSU task, the Cumulative style-fog-dual disentanglement Domain Adaptation method (CuDA-Net) is proposed in this paper. 
In this paper, we propose the Cumulative style-fog-dual disentanglement Domain Adaptation method (CuDA-Net) for the SFSU task.
We assume that the dual (style and fog) domain gap exists in SFSU, and that style, fog and dual factors have a cumulative relationship. 
Our method outperforms state-of-the-art methods on three widely-used datasets in SFSU and shows generalization ability to other adverse scenes, such as rainy and snowy scenes. 
We will make the code publicly available.
% Our methods outperforms state-of-the-arts on three widely used datasets in SFSU and shows generalization ability on other adverse scene, such as rainy and snowy scenes.
% the Foggy-Zurich, Foggy Driving and ACDC-fog datasets. 

\heading{Limitation.}
(1) %Selection of the baseline. 
Our method chose DISE~\cite{chang2019all} as our baseline, which can be replaced with other new stronger disentanglement-based domain adaptation methods. By doing so, we believe our CuDA-Net can achieve better performance.
(2) %More detailed analysis of the generalization ability. 
We conduct primary experiments to showcase certain generalization ability to rainy and snowy scenes, and more detailed analysis can be done to verify whether the cumulative relationship exists in other adverse settings.

% Very fortunately, our method also demonstrates the ability of defogging. 
% We hope and believe that our proposed domain gap decomposition for different factors are suitable for the domain adaptation task with multiple domain discrepancies.
% % To our knowledge, fog and style are intrinsically different. 
% In this paper, we propose a general framework to disentangle different factors, which can be applicable for other factors, such as rainy scenes and snowy scenes. 
% In the future, we would like to investigate more on the fog and style themselves and propose specific modules to benefit the disentanglement. 

% and fog synthesis, which are virtually better than related methods that require pair-wise training data.

%%%%%%%%% REFERENCES
{\small
\bibliographystyle{ieee_fullname}
\bibliography{egbib}
}

\end{document}